\documentclass[11pt]{article}

\usepackage[preprint]{acl}

\usepackage{times}
\usepackage{latexsym}

\usepackage[T1]{fontenc}

\usepackage[utf8]{inputenc}

\usepackage{microtype}

\usepackage{inconsolata}

\usepackage{graphicx}

\usepackage[table,xcdraw]{xcolor}
\usepackage{graphicx}
\usepackage{caption}      
\usepackage{subcaption}   
\usepackage{multirow}
\usepackage{booktabs} 
\usepackage{amsmath, amsthm, amssymb}
\usepackage{algorithm}
\usepackage{algpseudocode}
\usepackage{tabularx}

\usepackage{adjustbox}

\renewcommand{\thefootnote}{\fnsymbol{footnote}}

\title{QAgent: A modular Search Agent with Interactive Query Understanding}

\author{
 \textbf{Yi Jiang\footnotemark[1]\footnotemark[3]}, 
 \textbf{Lei Shen\footnotemark[1]}, 
 \textbf{Lujie Niu},
 \textbf{Sendong Zhao\footnotemark[2]},
 \textbf{Wenbo Su},
 \textbf{Bo Zheng}
\\
 Taobao \& Tmall Group of Alibaba
\\
 \texttt{jiangyijcx@163.com, zhaosendong@gmail.com}
}

\begin{document}
\maketitle

\footnotetext[1]{Equal contribution.}
\footnotetext[2]{Corresponding Author.}
\footnotetext[3]{Work done during an internship at Alibaba}

\renewcommand{\thefootnote}{\arabic{footnote}}

\begin{abstract}
Large language models (LLMs) excel at natural language tasks but are limited by their static parametric knowledge, especially in knowledge-intensive task. 
Retrieval-augmented generation (RAG) mitigates this by integrating external information. 
However, (1) traditional RAG struggles with complex query understanding, and (2) even search agents trained with reinforcement learning (RL), despite their promise, still face generalization and deployment challenges. 
To address these limitations, we propose QAgent, a unified agentic RAG framework that employs a search agent for adaptive retrieval. 
This agent optimizes its understanding of the query through interactive reasoning and retrieval.
To facilitate real-world application, we focus on modular search agent for query understanding that are plug-and-play in complex systems.
Secifically, the agent follows a multi-step decision process trained with RL to maximize retrieval quality and support accurate downstream answers. 
We further analyze the strengths and weaknesses of end-to-end RL and propose a strategy that focuses on effective retrieval, thereby enhancing generalization in LLM applications. 
Experiments show QAgent excels at QA and serves as a plug-and-play module for real-world deployment.
The code is available\footnote{https://github.com/OpenStellarTeam/QAgent.git .}.
\end{abstract}

\section{Introduction}

Large language models (LLMs) have demonstrated outstanding performance across a wide range of natural language processing tasks\citep{jaech2024openai,guo2025deepseek}. However, for complex, knowledge-intensive tasks, LLMs face significant challenges, including outdated knowledge and hallucinations\citep{bechard2024reducing,gade2025s}, which undermine their accuracy and reliability. 
Retrieval-augmented generation (RAG) enables LLMs to access external knowledge by retrieving documents and using them as context for generation, demonstrating great potential in addressing these limitations\citep{lewis2020retrieval,zhao2024retrieval1,fan2024survey,li2025matching}.

\begin{figure}
    \centering
    \includegraphics[width=0.825\linewidth]{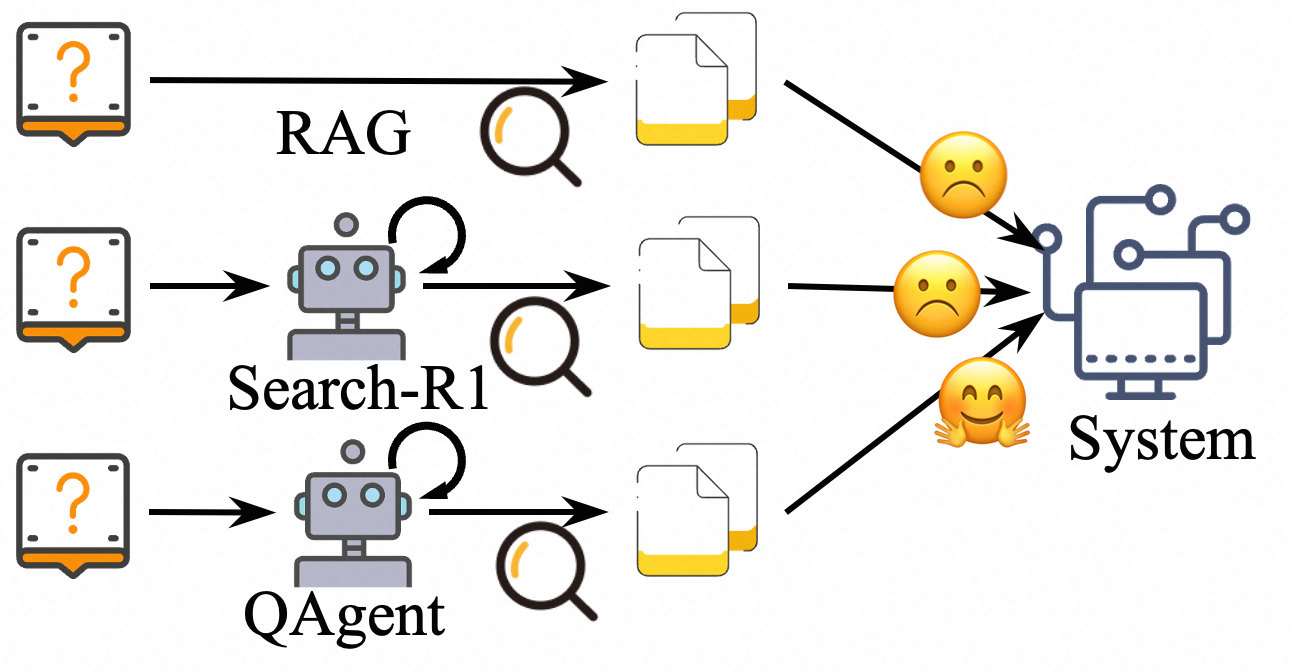}
    \caption{Different methods applied to the system. }
    \vspace{-1em}
    \label{fig:application}
\end{figure}

\begin{figure*}[!ht]
    \centering
    \includegraphics[width=0.975\linewidth]{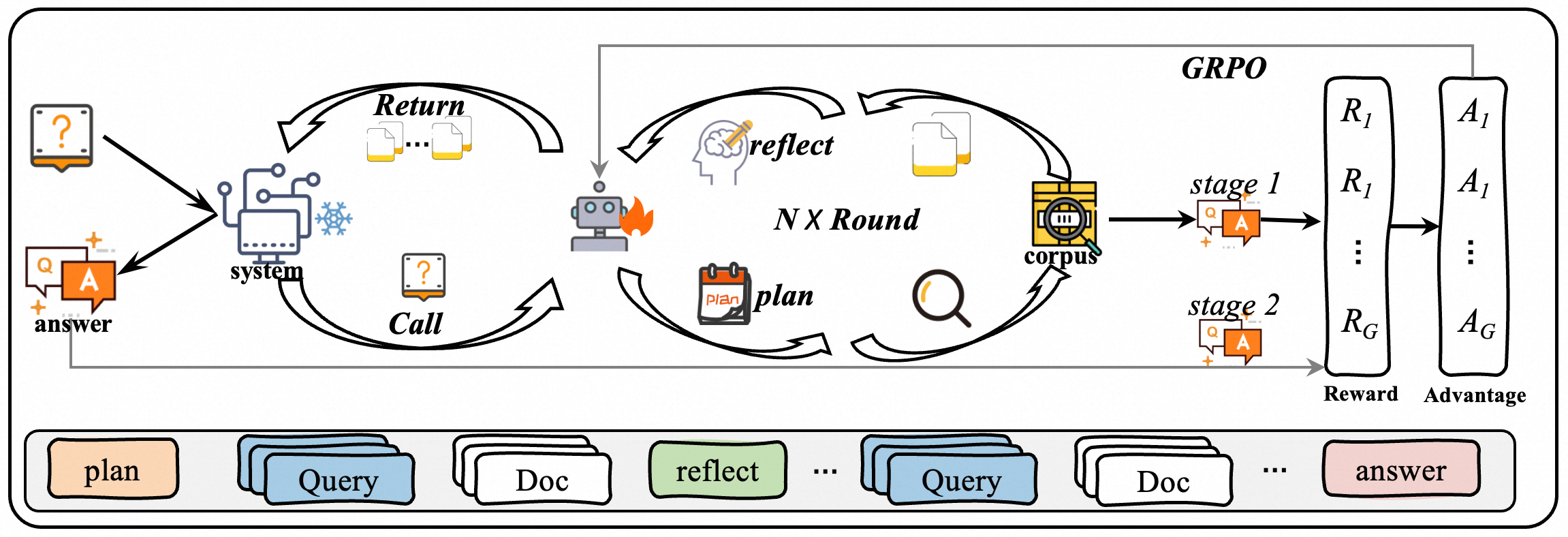}
    \caption{The proposed QAgent framework, with the left representing the system and the right representing our agent design, including ``plan-search-information-reflect''. In addition, the illustration shows a two-stage training framework, with the first stage using end-to-end RL and the second stage introducing generalized training. During training, the searched content tokens are masked to exclude them from loss calculations.}
    \label{fig:framework}
\end{figure*}

Nevertheless, the rigid workflow of traditional RAGs makes them inefficient for real-world problems requiring complex, multi-step reasoning. 
On the one hand, limitations exist in their ability to determine when and what to retrieve, making it difficult to construct accurate search paths for complex problems. 
On the other hand, RAGs are often embedded within complex systems, and the standalone RAG modules cannot adapt or optimize themselves according to the broader system’s needs. 
To alleviate these limitations, recent work has proposed various solutions, including query optimization\citep{ma2023query}, planning\citep{jiang2025retrieve}, reflection\citep{asai2024self}, active retrieval\citep{jeong2024adaptive}, and iterative retrieval\citep{trivedi2023interleaving}. 
While these approaches have increased flexibility and reduced supervision costs, they still fail to effectively leverage feedback for continuous iteration and performance optimization. DeepSeek-R1\citep{guo2025deepseek} demonstrates that even rule-based, outcome-driven rewards can train powerful reasoning agents. This has led to the emergence of search agent trained with RL\citep{jin2025search}, enabling autonomous reasoning, on-demand retrieval, and iterative query optimization.

In addition, practical deployment of search agents remains challenging. 
As shown in Figure~\ref{fig:application}, 
on the one hand, due to query complexity, the information directly retrieved from the original query may not be useful; on the other hand, due to the complexity of the system, the search agent is often used as a submodule, aiming to improve its information retrieval capability rather than its information utilization capabilities.

From this, we identify two key goals for practical search agents: 
(1) understand and decompose complex queries to bridge the gap between the original query and the retriever; 
and (2) retrieve information that benefits the downstream generation, ensuring generalizability when deployed as a submodule.

To address these challenges, we propose a unified framework that integrates query understanding, reasoning, and iterative refinement into a coherent search process. The framework enables adaptive query decomposition through multiple rounds of interaction, where the model progressively refines its understanding of complex user intents by generating, executing, and reflecting on retrieval actions. 
At the core of the framework is a lightweight search agent capable of autonomous reasoning and decision-making. 
By formulating search as a sequential decision-making problem, the agent is trained end-to-end using reinforcement learning to maximize the quality of information returned to the downstream generation system. 
This allows the agent to learn not only what to retrieve, 
but also how to re-optimize in a context-aware manner, ultimately acting as a submodule to improve overall system performance.

Our contributions are summarized as follows:
\begin{itemize}
    \item We propose a unified Agentic framework, centered around query understanding. 
    Through multiple rounds of interaction, QAgent unifies query optimization, and continues to evolve through outcome reward feedback.
    \item We analyze the strengths and weaknesses of end-to-end reinforcement learning training and propose a two-stage training solution, providing profound insights for the practical application of search agents.
    \item We conduct extensive experiments demonstrate that QAgent delivers strong performance. And further analysis confirms its meaningful gains in information composition.
\end{itemize}

\section{Related Work}
\subsection{Retrieval-Augmented Generation}
Retrieval-augmented generation (RAG) has emerged as a promising approach to mitigate limitations of the LLMs, such as hallucination\citep{bechard2024reducing,gade2025s}, by augmenting them with external retrieved knowledge\citep{lewis2020retrieval,zhao2024retrieval1,fan2024survey,li2025matching,zhou2024trustworthiness}. 
Classic RAG systems typically follow a ``retrieve-then-read’’ paradigm\citep{lewis2020retrieval}, 
but it still faces numerous challenges that hinder optimal performance. 
To handle the complexity of problem, recent work identifies retrieval intent to skip unnecessary retrieval \citep{jeong2024adaptive,cheng2024unified,wang2023self} and refines queries to better match retriever expectations \citep{ma2023query,wang2023query2doc,xu2024search}. To counter noisy or irrelevant passages, methods introduce denoising\citep{wei2025instructrag}, re-ranking\citep{glass2022re2g,xiao2024c}, and compression\citep{xu2024recomp,jiang2023longllmlingua} between retrieval and generation. To bridge the retriever-generator gap, some jointly optimize both modules \citep{izacard2023atlas,lin2023ra} or insert trainable middlewares \citep{jiang2025gainrag,ke2024bridging,dong2025understand}.
Recent progress has increasingly moved towards agentic RAGs: Self-RAG\citep{asai2024self} combines on-demand retrieval with reflection, MetaRAG\citep{zhou2024metacognitive} iteratively acquires knowledge through metacognition, and Search-o1\citep{li2025search} performs retrieval while reasoning. 
However, current RAG systems exhibit limited agentic capabilities and cannot evolve continuously, due to the lack of effective optimization methods.

\subsection{Search Agent with RL}
Reinforcement learning (RL)\citep{kaelbling1996reinforcement} has emerged as a promising paradigm for improving the reasoning capabilities of large language models (LLMs). By interacting with the environment and driven by the goal of maximizing rewards, models are able to achieve autonomous reasoning and decision-making, such as  OpenAI-o1\citep{jaech2024openai} and DeepSeek-R1\citep{guo2025deepseek}.
Recent research has further integrated RL into agent-based RAGs, enabling search agents to continuously learn and evolve through real-time interactions with search engines. For example, Search-R1\citep{jin2025search}, R1-Searcher\citep{song2025r1}, Deepresearcher\cite{zheng2025deepresearcher} and ZeroSearch\citep{sun2025zerosearch} utilize RL to train models for simultaneous search and reasoning, eliminating the need for supervision over intermediate reasoning steps. 
Subsequently, many improvements were made in reasoning patterns\citep{ren2025effective}, reward verification\citep{wang2025stepsearch,he2025sufficiency} and efficiency\citep{jiang2025s3,sha2025sem}.
While these approaches have achieved impressive performance, their practical applicability remains unclear due to limitations in efficiency and generalization. 
In contrast, our work focuses on query understanding and analyzes training strategies that promote generalization, providing a solution for integrating search agents into complex systems in real-world applications. 

\section{Methodology}
\label{Method}
In this section, we introduce our agentic framework shown in Figure~\ref{fig:framework}. It has two parts: the inference loop and the training strategy. 
The inference loop introduces our multi-round rollout for search agent.
The training strategy first analyzes the RL training and then introduces a two-stage training. 

\subsection{Preliminaries}
\subsubsection{Agentic RAG}
\label{sec:preliminaries}
Naive RAG follows a retrieve-read paradigm. 
Let $q$ be a query and $\mathcal{D}$ a document corpus. A retrieval function (including a retriever and a corpus) returns $k$ relevant documents, denoted $\mathcal{R}(q)$. A generator $\mathcal{G}$ produces a answer $\hat{A}$ conditioned on $q$ and $\mathcal{R}(q)$:

\begin{equation}
    \hat{A} = \mathcal{G}(q, \mathcal{R}(q)).
\end{equation}
In the agentic setting, search is modeled as a sequential process governed by policy $\pi_\theta(a_t\mid s_{<t})$, where $s_{<t}$ includes $q$, past actions, retrieved documents, and reasoning history. The agent selects actions $a_t \in \mathcal{A}$:
\begin{equation}
a_t \sim \pi_\theta(\cdot \mid s_{<t}),
\end{equation} 
typically reasoning or retrieval. 
The interaction yields a trajectory $\tau = (s_0, a_0, \dots, s_T, a_T)$, with a final answer, framing RAG as a multi-step decision process for dynamic reasoning and retrieval.

\subsubsection{Reinforcement Learning}
\label{sec:RL}
Reinforcement learning (RL) is used to train the policy $\pi_\theta(y|x)$ by maximizing task-specific rewards $r_\phi(x, y)$. We adopt Group Relative Policy Optimization (GRPO)\citep{shao2024deepseekmath}, which operates on $G$ rollouts $\{y_i\}_{i=1}^G$ to compute group-relative advantages. The objective is:
\begin{equation}
\begin{aligned}
    \mathcal{J}(\theta) &= \mathbb{E}_{x \sim \mathcal{D},\, \{y_i\}} \big[ \frac{1}{\sum_{i=1}^{G} |y_i|} \sum_{i=1}^{G} \sum_{t=1}^{|y_i|} 
    \min \big( \\
    & \rho_{i,t} A_{i},\, \text{clip}\left( \rho_{i,t}, 1 - \epsilon, 1 + \epsilon \right) A_{i} \big) - \beta \mathbb{D}_{\text{KL}}
    \big],
\end{aligned}
\end{equation}
where $\rho_{i,t} = \pi_\theta(y_{i,t}|x, y_{i,<t}) / \pi_{\theta_{\text{old}}}(y_{i,t}|x, y_{i,<t})$, and $A_i = (R_i - \mu_R) / \sigma_R$ is the normalized advantage with $\mu_R$ and $\sigma_R$ the mean and standard deviation of $\{R_j\}_{j=1}^G$. 
The KL term 
$\mathbb{D}_{\text{KL}} = \mathbb{D}_{\text{KL}}\left( \pi_\theta \parallel \pi_{\theta_{\text{ref}}} \right)$ 
constrains policy deviation. 
Our training is based on this algorithm.

\subsection{Multi-Turn Query Optimation Loop}
\label{sec:mqol}
\begin{figure}[!ht]
    \centering
    \includegraphics[width=0.95\linewidth]{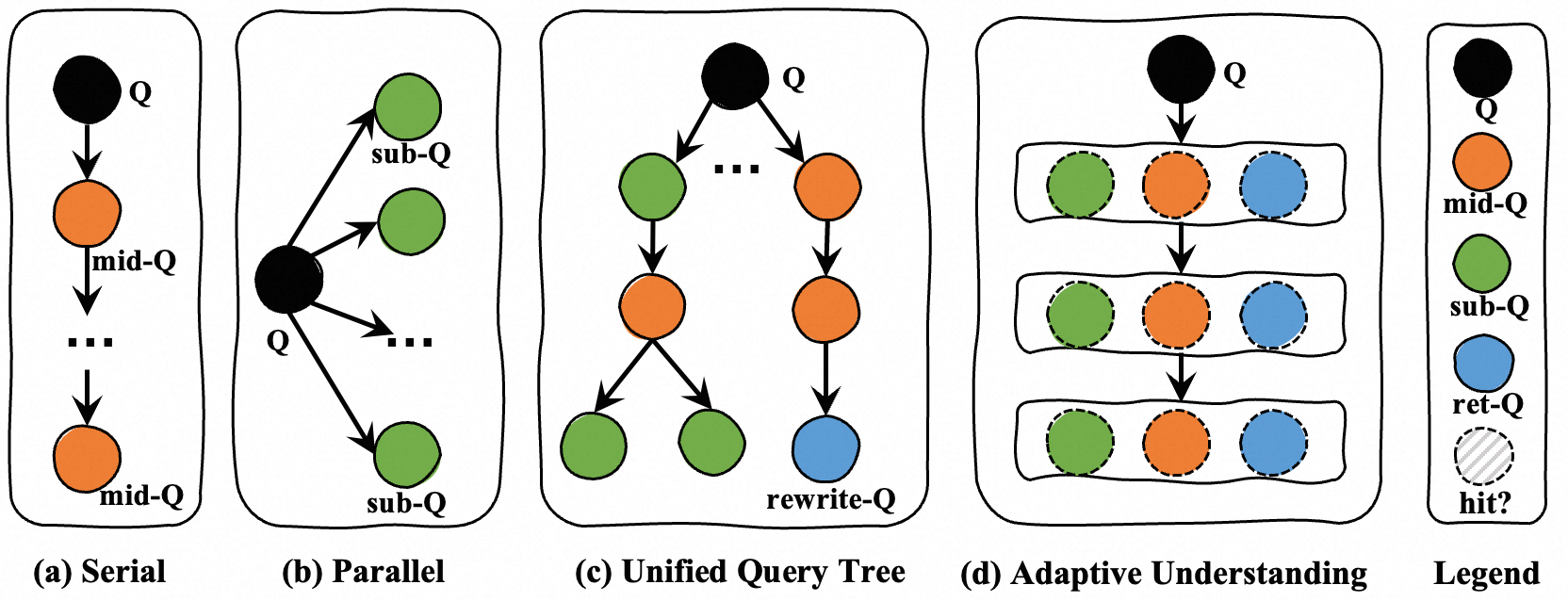}
    \caption{Illustration of query understanding. }
    \label{fig:query}
\end{figure}

\paragraph{Motivation.} 
One main goal of the search agent is to retrieve useful information. When the retriever and corpus are fixed, the way to achieve it is to optimize the query. 
Understanding the original query is complex\citep{zhao2024retrieval2}, including but not limited to: the query requires multi-hop reasoning (Fig.~\ref{fig:query}(a)), needs to be decomposed (Fig.~\ref{fig:query}(b)), and does not conform to the retrieval system's preferences. 
Fully modeling this yields a wide and deep tree (Fig.~\ref{fig:query}(c)), which demands numerous model calls and retries, resulting in high complexity.

\paragraph{Workflow.}
To address these challenges, inspired by recent advances in multi-round agentic RAG, we model the expansion of query understanding as a stochastic sequential decision process (Figure~\ref{fig:query}(d)). Instead of enforcing rigid transformation rules, our framework enables the agent to autonomously decide how to refine and issue queries over multiple interactions with a retrieval system. 

Formally, given an initial query $q$ , and conditioned on a fixed retriever $\mathcal{R}$ and retrieval corpus $\mathcal{D}$, the agent generates a sequence of optimized queries $S = \{S_1, \dots, S_T\} $ over $ T $ rounds, where each $ S_t = \{q_{t,1}, \dots, q_{t,m_t}\} $ is a set of transformed queries produced at round $ t $. The full trajectory probability is factorized (see Figure~\ref{fig:framework}):
\begin{equation}
\adjustbox{width=0.975\linewidth, valign=c}{
$\displaystyle
P(S \mid q, \mathcal{R}, \mathcal{D}) = \prod_{t=1}^{T} P\left(S_t \mid q, ( I^{\text{pre}}_{i},\, S_{i},\, \mathcal{C}_{i},\, I^{\text{post}}_{i} )_{i<t} \right)
$%
}
\label{eq:query_generation}
\end{equation}
\medskip
where $(\cdot)_{i<t}$ donates historical actions before round t, and $ I^{pre}_{i} $, $ I^{post}_{i} $ encode pre-retrieval and post-retrieval reasoning traces (e.g., plan and reflection).
The $C_i$ donate the retrieval context, which aggregates results from all transformed queries:
\begin{equation}
C_i = \oplus_{j=1}^{m_i} \mathcal{R}(q_{ij}), \quad q_{ij} \in S_i,
\label{eq:ci}
\end{equation}
where $ \mathcal{R}(\cdot) $ fetches relevant documents from $ \mathcal{D}$, and $ \oplus $ represents context aggregation. 

Overall, the agent follows a structured multi-round interaction loop, formalized as a trajectory: 
\begin{equation}
\tau = (q, I^{pre}_1, S_1, \mathcal{C}_1, I^{post}_1, \dots, \mathcal{C}_{T}, I^{post}_{T}, \hat{A} ).
\label{eq:interaction_loop}
\end{equation}
At each round $ t $, the agent performs planning $ I^{pre}_t $, generates optimized queries $ S_t $. Followed by reflection $ I^{post}_t $, in which the agent assesses the completeness of the accumulated information. 

By unifying multiple query transformation paradigms within a stochastic, reflective interaction loop, our framework supports richer and more flexible query understanding than methods limited to fixed decomposition patterns, enabling the model to \textit{ask better questions}.

\subsection{End-to-End RL Training}
For optimizing search agent, reinforcement learning is a powerful tool. It enables continuous interaction with the environment and uses outcome-based rewards for end-to-end optimization, thereby enhancing the LLM’s agentic capability.

\paragraph{Reward Design.}
In the end-to-end RL training, our rewards are mainly designed based on two factors: the correctness of the answer and the correctness of the format. Formally, 
\begin{equation}
    R(\tau) = \mathbb{I}\{r_{\mathrm{fmt}}(\tau) = 1\} \cdot \mathrm{EM_{s}}(A^*, \hat{A}). 
\end{equation}
$\mathrm{EM_{s}}$ stands for Strict EM, the same as Search-R1. The reward format supports pre-retrieval planning, post-retrieval reflection, and final answer tags.
This strict, sparse design may waste rollouts in early training, but ensures rollout quality and gradually improves efficiency as training stabilizes.

\subsection{Generalized RL Training}
\label{sec:Generalized_Train}
\paragraph{Motivation.} 
End-to-end training significantly improves the agent's capabilities, both in information retrieval and utilization. This training optimizes both retrieval and generation. 
However, due to the high latency of current autoregressive LLMs and the complexity of real-world systems, a single monolithic model is often insufficient to fully handle all tasks. We envision a search agent as a component that prioritizes information retrieval, while information utilization is not necessarily important. 
In this context, query optimization and information retrieval are our primary goals, and improving information utilization during training can become a form of reward hacking. 

\begin{figure}
    \centering
    \includegraphics[width=0.975\linewidth]{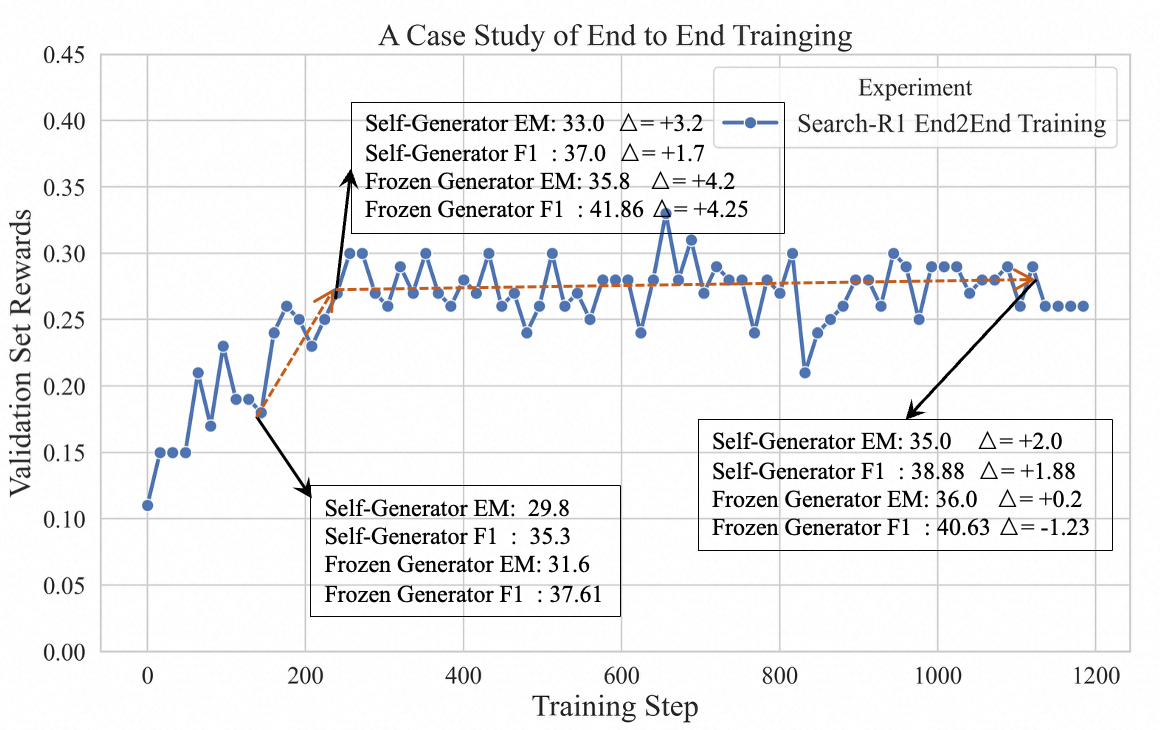}
    \caption{A case study of Agentic RL training (Search-R1), where both the fast ascent phase and the slow convergence phase are tested.}
    \label{fig:caseR1}
\end{figure}

\paragraph{Analysis.} 
We analyze a case study of the training process of an agent using reinforcement learning. As shown in Figure~\ref{fig:caseR1}, in the early stages of model training, information retrieval and information utilization mutually enhance each other. However, later in training, the model tends to prioritize information utilization to obtain rewards. This undoubtedly compromises the generalization ability of the search agent as a submodule. 
An analysis of information utilization can be seen in \S~\ref{sec:utilization}.

\paragraph{Generalized as Submodule.} 
To address this, we introduced the second training stage. This involves using a frozen generator to answer based on the agent’s retrieved documents, and then calculating rewards for responses from the frozen generator, thereby improving the agent’s generalization as a tool. 
Formally, \begin{align}
    \mathcal{K} &= \text{parseDocSet}(\tau), \\
    \tilde{A}   &= \mathcal{G}(q, \mathcal{K})
\end{align}
where $\mathcal{K}$ is the retrieved passage set and $\tilde{A}$ is the frozen generator’s predicted answer.

\paragraph{Reward Design.} 
In this stage, constraints are shifted to the correctness of the generator's response, with format and mandatory answering requirements relaxed.
\begin{equation}
    R(\tau) = \mathrm{EM}(A^*, \tilde{A})+0.5*Hit(\tau, A^*)
\end{equation}
$\mathrm{EM}$ is relaxed to non-strict EM because using strict EM on a frozen generator is inefficient, unlike a generator trained with a format reward.
In addition, the reward for the model's end-to-end answer will be relaxed to $Hit$, meaning that the entire trajectory contains the gold answer. 

\begin{algorithm}[!ht]
\caption{Rollout with Multi-Turn Interaction}
\label{alg:rollout}
\begin{algorithmic}[1]
\Require Input query $q$, policy model $\pi_\theta$, search engine $\mathcal{R}$, maximum turns $B$, Generator $\mathcal{G}$.
\Ensure Final predicted answer $\hat{A}$.
\State Initialization: rollout sequence $y \leftarrow \emptyset$, action count $b \leftarrow 0$, info set $\mathcal{K} \leftarrow \emptyset$
\While{$b < B$}
    \State $y_b \leftarrow \emptyset$ \Comment{Initialize current sequence}

    \Repeat
        \State $y_t \sim \pi_\theta(\cdot \mid q, y + y_b)$, \quad $y_b \leftarrow y_b \oplus y_t$
    \Until{\textless\text{/search}\textgreater , \textless\text{/answer}\textgreater , \textless\text{eos}\textgreater  in $y_b$}

    \State $y \leftarrow y + y_b$
    \If{\textit{SearchTags} detected in $y_b$}
        \State $Q \leftarrow \text{ParseQ}(S,$\textless\text{query}\textgreater , \textless\text{/query}\textgreater $)$
        
        \State $C = [\mathcal{R}(q_{1}), \cdots, \mathcal{R}(q_n)], q_i \in Q$
        
        \State $y\leftarrow y \oplus $\textless\text{information}\textgreater  $C$ \textless\text{/information}\textgreater
        \State $\mathcal{K} \leftarrow \mathcal{K} \cup C$ \Comment{Update info set}

    \ElsIf{\textit{AnswerTags} in $y_b$}
        \State \textbf{break}
    \Else
        \State Ask for retry: $y \leftarrow y \oplus prompt_{\text{rethink}}$ 
    \EndIf
    \State Increment action count $b \leftarrow b + 1$
\EndWhile
\State $\tilde{A} = \mathcal{G}(q,\mathcal{K})$\Comment{Generate final answer}
\State \textbf{return} final predicted answer $\tilde{A}$
\end{algorithmic}
\end{algorithm}

\begin{table*}[ht]
\begin{center}
\resizebox{0.925\linewidth}{!}{
\begin{tabular}{lllllllllllll}
\toprule
\multicolumn{1}{c}{\multirow{2}{*}{Method}} & \multicolumn{2}{c}{HotpotQA}   & \multicolumn{2}{c}{2WikiMHQ$^\dagger$}    & \multicolumn{2}{c}{MuSique$^\dagger$}     & \multicolumn{2}{c}{NaturalQA}  & \multicolumn{2}{c}{WebQ$^\dagger$}                 & \multicolumn{2}{c}{Average}                     \\
\cmidrule(lr){2-3} \cmidrule(lr){4-5} \cmidrule(lr){6-7} \cmidrule(lr){8-9} \cmidrule(lr){10-11}\cmidrule(lr){12-13}

\multicolumn{1}{c}{}        & \multicolumn{1}{c}{EM} & \multicolumn{1}{c}{F1} & \multicolumn{1}{c}{EM} & \multicolumn{1}{c}{F1} & \multicolumn{1}{c}{EM} & \multicolumn{1}{c}{F1} & \multicolumn{1}{c}{EM} & \multicolumn{1}{c}{F1} & \multicolumn{1}{c}{EM} & \multicolumn{1}{c}{F1} & \multicolumn{1}{c}{EM} & \multicolumn{1}{c}{F1} \\
\midrule
Search-o1   & 23.20   & 27.04  & 22.40   & 22.74   & 5.80     & 10.44  & 23.63  & 23.80  & 30.02  & 26.99   & 21.01 & 22.20 \\
Search-R1   & \underline{35.00}   & \underline{38.88}  & \underline{35.40}   & \underline{31.86}   & \underline{7.60}     & \textbf{14.01}  & \underline{30.89}  & \underline{32.37}  & \textbf{35.68}  & \underline{35.72}   & \underline{28.91} & \underline{30.57} \\
ZeroSearch  & 29.40   & 28.46  & 27.80   & 22.03   & 3.00     & 8.00   & 17.95  & 10.62  & 26.48  & 23.94   & 20.93 & 18.61 \\
QAgent    & \textbf{37.60}   & \textbf{44.44 } & \textbf{38.20}   & \textbf{36.11}   & \textbf{7.80}     & \underline{12.39}  & \textbf{31.47}  & \textbf{35.19}  & \underline{32.09}  & \textbf{38.01}   & \textbf{29.43} & \textbf{33.23} \\
\bottomrule
\end{tabular}
}
\end{center}
\caption{Main results of end-to-end performance. Best and second-best results are highlighted in \textbf{bold} and \underline{underlined}, respectively. $^\dagger$ denotes out-of-distribution evaluation datasets.}
\label{tab:results-E2E}
\end{table*}

\begin{table*}[h!t]
\begin{center}
\resizebox{0.925\linewidth}{!}{
\begin{tabular}{lllllllllllll}
\toprule
\multicolumn{1}{c}{\multirow{2}{*}{Method}} & \multicolumn{2}{c}{HotpotQA}   & \multicolumn{2}{c}{2WikiMHQ$^\dagger$}    & \multicolumn{2}{c}{MuSique$^\dagger$}     & \multicolumn{2}{c}{NaturalQA}  & \multicolumn{2}{c}{WebQ$^\dagger$}                 & \multicolumn{2}{c}{Average}                     \\
\cmidrule(lr){2-3} \cmidrule(lr){4-5} \cmidrule(lr){6-7} \cmidrule(lr){8-9} \cmidrule(lr){10-11}\cmidrule(lr){12-13}
\multicolumn{1}{c}{}        & \multicolumn{1}{c}{EM} & \multicolumn{1}{c}{F1} & \multicolumn{1}{c}{EM} & \multicolumn{1}{c}{F1} & \multicolumn{1}{c}{EM} & \multicolumn{1}{c}{F1} & \multicolumn{1}{c}{EM} & \multicolumn{1}{c}{F1} & \multicolumn{1}{c}{EM} & \multicolumn{1}{c}{F1} & \multicolumn{1}{c}{EM} & \multicolumn{1}{c}{F1} \\
\midrule
\multicolumn{13}{c}{\cellcolor[HTML]{EFEFEF}Generator 3B} \\        

Vanilla     & 15.00  & 19.33  & 23.00  & 25.01  & 0.80   & 6.40   & 17.53  & 19.58  & 28.44  & 29.38  & 16.95  & 19.94                  \\
NaiveRAG    & 30.60  & 35.54  & 25.40  & 28.32  & 3.20   & 7.64   & 30.22  & 31.43  & 29.97  & 30.26  & 23.88  & 26.64                  \\
Rerank      & \underline{34.40}  & \underline{39.87}  & 28.40  & \underline{30.95}  & 3.80   & 9.49   & \textbf{35.35}  & \textbf{35.20}  & \underline{32.97}  & \underline{31.92}  & \underline{26.98}  & \underline{29.48}       \\
Search-o1   & 21.20  & 25.14  & 24.60  & 26.02  & 2.20   & 7.33   & 24.07  & 24.39  & 32.08  & 30.66  & 20.83  & 22.71                  \\
Search-R1   & 31.80  & 36.71  & \underline{29.40}  & 30.83  & \underline{5.20}   & \underline{10.14}  & 28.39  & 28.86  & 28.39  & 28.86  & 24.64  & 27.08                  \\
zero-search & 26.40  & 30.46  & 24.80  & 26.07  & 2.00   & 4.98   & 19.31  & 20.40  & 23.13  & 24.74  & 21.17  & 24.86  \\
QAgent  & \textbf{37.00}  & \textbf{42.94}  & \textbf{33.80}  & \textbf{35.22}  & \textbf{5.20}   & \textbf{10.20}  & \underline{33.82}  & \underline{33.68}  & \textbf{36.32}  & \textbf{32.78}  & \textbf{29.23}  & \textbf{30.96}  \\
\midrule
\multicolumn{13}{c}{\cellcolor[HTML]{EFEFEF}Generator 7B} \\                                                                                                                                                                                                                    
Vanilla     & 20.20  & 25.54  & 23.40  & 27.00  & 4.80   & 10.77  & 22.80  & 25.72  & 32.73  & 35.87  & 20.79  & 24.98                  \\
NaiveRAG    & 36.00  & 42.47  & 28.00  & 31.24  & 4.40   & 10.78  & 33.30  & 35.21  & 31.55  & 33.52  & 26.65  & 30.64                  \\
Rerank      & \underline{39.00}  & \underline{44.76}  & 28.40  & 32.11  & 5.40   & 11.84  & \textbf{38.28}  & \textbf{39.06}  & 33.66  & 34.75  & \underline{28.95}  & \underline{32.50}                  \\
Search-o1   & 27.00  & 31.15  & 24.40  & 27.78  & 6.00   & 11.77  & 28.42  & 30.06  & \underline{33.91}  & \underline{35.40}  & 23.95  & 27.23                  \\
Search-R1   & 36.00  & 40.63  & \underline{34.40}  & \underline{33.41}  & \underline{7.00}   & \underline{12.35}  & 31.27  & 32.09  & 32.07  & 33.94  & 28.15  & 30.48                  \\
ZeroSearch  & 28.40  & 34.44  & 24.00  & 27.10  & 3.00   & 7.47   & 23.21  & 25.00  & 27.26  & 30.27  & 21.17  & 24.86                  \\
QAgent  & \textbf{42.40}  & \textbf{46.75}  & \textbf{35.80}  & \textbf{36.76}  & \textbf{7.40}   & \textbf{13.81}  & \underline{37.37}  & \underline{38.87} & \textbf{36.71}  & \textbf{36.41}  & \textbf{31.94}  & \textbf{34.52} 
\\ \bottomrule
\end{tabular}
}
\end{center}
\caption{Main results when used as a submodule. Best and second-best results are highlighted in \textbf{bold} and \underline{underlined}, respectively. $^\dagger$ denotes out-of-distribution evaluation datasets.}
\label{tab:results-frozen}
\end{table*}

\subsection{Inference}
For each query, QAgent starts with task instructions and reasons step by step. When it hits  \textless search\textgreater...\textless/search\textgreater tags, it extracts the query, retrieves documents, and appends them wrapped in \textless information\textgreater...\textless/information\textgreater tags. This loop repeats, building a reasoning chain with external evidence. Finally, the passage set is fed to the generator to produce the answer (see Algorithm~\ref{alg:rollout}).


\section{Experiment}
\label{Experiment}
This section presents the experimental results, and analysis. We main address the following questions:
RQ1: How does QAgent perform in end-to-end QA?(\S~\ref{sec:main_results})
RQ2: How well does QAgent function as a submodule?(\S~\ref{sec:main_results})
RQ3: How does RL training affect information retrieval or usage?(\S~\ref{sec:main_results} and \S~\ref{sec:utilization})
RQ4: What advantages does the QAgent paradigm offer over traditional RAG?(\S~\ref{sec:Gain})

\subsection{Experimental Setup}
\paragraph{Datasets.} We evaluate QAgent on five open-domain QA datasets covering both multi-hop and single-hop reasoning. 
The multi-hop QA benchmarks include \textbf{2WikiMultiHopQA}\citep{ho2020constructing}, \textbf{HotpotQA}\citep{yang2018hotpotqa} and \textbf{Musique}\citep{trivedi2022musique}. 
For general QA, we use \textbf{WebQuestions}\citep{berant2013semantic} and \textbf{NaturalQA}\citep{kwiatkowski2019natural}.
We adopt 500-sample subsets of 2WikiMultiHopQA and HotpotQA for efficiency, following \citet{trivedi2023interleaving}. More details are provided in Appendix~\ref{appx:dataset}.

\begin{table*}[!htbp]
\centering
\resizebox{0.925\linewidth}{!}{
\begin{tabular}{rllllllllllll}
\toprule
\multicolumn{1}{c}{\multirow{2}{*}{Method}} & \multicolumn{2}{c}{HotpotQA}   & \multicolumn{2}{c}{2WikiMHQ$^\dagger$}    & \multicolumn{2}{c}{MuSique$^\dagger$}     & \multicolumn{2}{c}{NQ}  & \multicolumn{2}{c}{WebQ$^\dagger$}                 & \multicolumn{2}{c}{Average}                     \\
\cmidrule(lr){2-3} \cmidrule(lr){4-5} \cmidrule(lr){6-7} \cmidrule(lr){8-9} \cmidrule(lr){10-11}\cmidrule(lr){12-13}
\multicolumn{1}{c}{}        & \multicolumn{1}{c}{EM} & \multicolumn{1}{c}{F1} & \multicolumn{1}{c}{EM} & \multicolumn{1}{c}{F1} & \multicolumn{1}{c}{EM} & \multicolumn{1}{c}{F1} & \multicolumn{1}{c}{EM} & \multicolumn{1}{c}{F1} & \multicolumn{1}{c}{EM} & \multicolumn{1}{c}{F1} & \multicolumn{1}{c}{EM} & \multicolumn{1}{c}{F1} \\
\midrule
\multicolumn{13}{c}{\cellcolor[HTML]{EFEFEF}Generator 3B} \\                    
QAgent   w/o all          & 31.20  & 37.03  & 29.20  & 30.81  & 4.40   & 9.67   & 28.95  & 29.63  & 30.36  & 29.87  & 24.82  & 27.40 \\
+stage1                     & 36.20  & 41.21  & 30.60  & 33.19  & \textbf{5.40}   & \textbf{10.24}  & 30.03  & 30.04  & 30.91  & 30.20  & 26.63  & 28.98 \\
+stage1+stage2              & \textbf{37.00}  & \textbf{42.94}  & \textbf{33.80}  & \textbf{35.22}  & 5.20   & 10.20  & \textbf{33.82}  & \textbf{33.68}  & \textbf{36.32}  & \textbf{32.78}  & \textbf{29.23}  & \textbf{30.96} \\
\midrule
\multicolumn{13}{c}{\cellcolor[HTML]{EFEFEF}Generator 7B} \\ 
QAgent   w/o all          & 36.00  & 41.84  & 29.80  & 32.64  & 5.40   & 11.32  & 31.75  & 33.63  & 31.79  & 33.51  & 26.95  & 30.59 \\
+stage1                     & 40.40  & 45.84  & 35.20  & 37.65  & 7.40   & 13.60  & 33.52  & 34.93  & 32.58  & 33.83  & 29.82  & 33.17 \\
+stage1+stage2              & \textbf{42.40}  & \textbf{46.75}  & \textbf{35.80}  & \textbf{36.76}  & \textbf{7.40}   & \textbf{13.81}  & \textbf{37.37}  & \textbf{38.87}  & \textbf{36.71}  & \textbf{36.41}  & \textbf{31.94}  & \textbf{34.52}   
\\ \bottomrule
\end{tabular}
}
\caption{Ablation Study. Best results are highlighted in \textbf{bold}. $^\dagger$ denotes out-of-distribution evaluation datasets.}
\end{table*}

\paragraph{Evaluation.} We report two metrics: EM and F1. Following prior work\citep{asai2024self}, we use a non-strict EM where a prediction is correct if it contains the gold answer. F1 measures token-level overlap between the predicted and gold answers. 
Since longer response may improve EM via coverage but introduce noise that lowers F1, evaluating both metrics allows for a more balanced evaluation.

\paragraph{Baselines.} We compare QAgent with different baselines: 
(1) Vanilla and Naive RAG. Vanilla: direct answering without retrieval and Naive RAG: answering in the ``retrieval-reading'' paradigm. 
(2) Search-Agent. We utilize Search-o1, which uses multiple rounds of interaction but without RL training, and Search-R1 and zero-search, which use multiple rounds of interaction and joint retrieval and generation with RL training.
(3) As the search agent component. We extract the retrieved documents from the model inference trajectory and input them into the frozen generators: Search-o1, Search-R1, zero-search, and the middleware Rerank.
Details are in Appendix~\ref{appx:baselines}
\paragraph{Implementation Details.} We use Qwen-2.5-3B (Instruct)\citep{qwen2025qwen25technicalreport} as the training model and use frozen as the second-stage supervision model. 
For generation, we use Qwen-2.5-3B (Instruct) and 7B as generators. For retrieval, we use the 2018 Wikipedia dump as the knowledge source, and all baselines use bm25\citep{robertson2009probabilistic} as the retriever. 
For fairness, we retrieve 1 passage each query and deduplicate before sending to the generator. 
And the reported Search-R1 is a strictly aligned reproduction based on our framework. 

\subsection{Main Results}
\label{sec:main_results}
We present comprehensive evaluation results in Table~\ref{tab:results-E2E} and Table~\ref{tab:results-frozen}, respectively. From the results, we have the following observations:

\paragraph{End-to-End QA performance.} 
Overall, the RL-trained methods outperform the training-free methods, demonstrating the potential of RL training (the Zerosearch evaluation was not as good as expected, which we analyzed may be due to the sensitivity of the search engine). 
In an end-to-end QA comparison, our training achieved an improvement of 0.52\% and 2.66\% in EM and F1 respectively compared to Search-R1. 
Furthermore, we observed that on multi-hop datasets like Musique, our approach did not achieve significant improvement compared to Search-R1. We speculate that this may be due to our adaptive query optimization pattern. While it expands both the depth and breadth of query understanding, the inevitable expansion of breadth slightly compromises the depth.

\paragraph{Performance as Submodule.} 
Table~\ref{tab:results-frozen} shows the generalization of the search agent as a submodule in complex systems.
We found that when freezing 3B as the generator, the overall performance of Search-R1 was significantly lower than that of end-to-end QA, indicating that its generalization was somewhat insufficient. 
In contrast, our method maintains strong generalization. 
Specifically, our approach improves average EM by 5.35\% over NaiveRAG.
It also outperforms end-to-end optimized search-R1 by 4.59\% in average EM.
And we found that the middleware rerank still achieves relatively good performance, but it is a non-conflicting submodule with QAgent, which means they can be used simultaneously. 
Moreover, integrating our QAgent (3B) into a system with a 7B generator yields even greater gains, suggesting that complex systems can rely on a strong generator equipped with small search agent as submodule.

\subsection{Ablation Study}
In this section, we conduct detailed ablation experiments to verify the effectiveness of our training at each stage. 
As shown in the figure, our train-free version achieves high performance, with a significant improvement after end-to-end training. This benefited from the reinforcement learning algorithm, but end-to-end training still has shortcomings. We found that it significantly improved the performance in in-distribution evaluation, but the improvement was limited in out-of-distribution evaluation. 
After introducing a second stage of training, the performance was further improved, especially due to its improved generalization.

\begin{figure}[!h]
    \centering
    \includegraphics[width=0.975\linewidth]{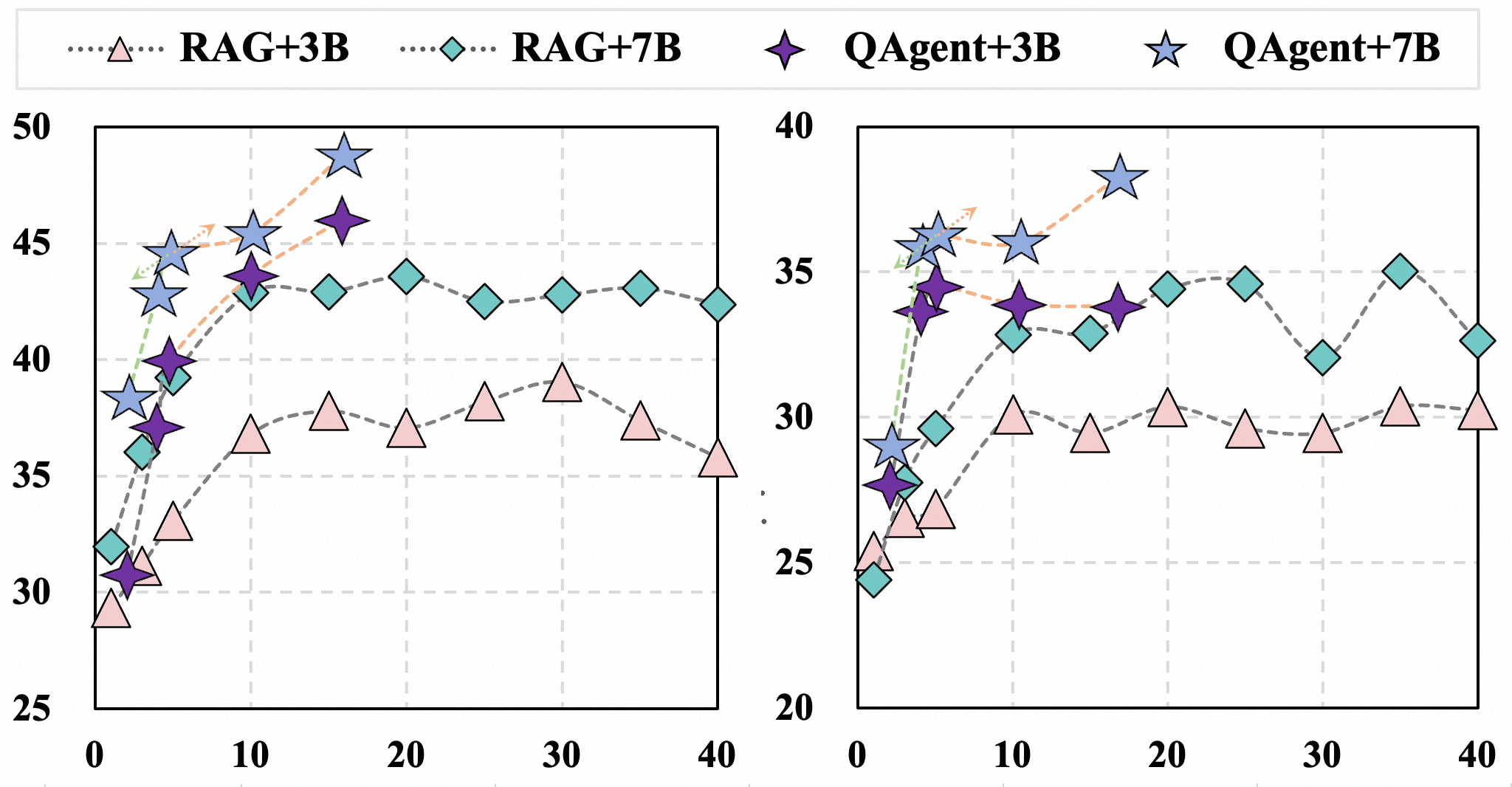}
    \caption{Illustration of the combined gain. The left is the experimental results of HotpotQA, and the right is that of 2WikiMHQ.}
    \label{fig:gain}
\end{figure}
\subsection{Combined Gain Analysis}
\label{sec:Gain}
In this section, we analyzed the advantages of using the search agent as a submodule. 
Specifically, we increased the top-k threshold under Naive RAG to explore the upper limits of the ``retrieval-reading'' paradigm. As shown in Figure~\ref{fig:gain}, due to the complexity of the problem, even with an increased top-k threshold, the gains gradually diminished. This is likely due to the redundancy of candidate document information and the limited amount of combinatorial information, which is an unavoidable limitation of all retrievers. 
The search agent, acting as an middleware between the query and the retriever, addresses the limitations of the retriever. Using the same token budget as the generator, it achieves even greater ``combination gains'', even exceeding the upper limits of the traditional paradigm.

Furthermore, we verified that using only the first few rounds significantly outperformed NaiveRAG, demonstrating its effectiveness.
We also increased the retrieved top-k per query and found that performance continued to improve, but the number of documents used also increased significantly.

Overall, QAgent increases the upper limit of the model's performance when handling complex problems. However, in practical applications, a trade-off between the number of documents used and efficiency is necessary.

\begin{figure}[!tp]
  \centering
    \begin{minipage}[b]{0.45\linewidth}
        \centering
        \includegraphics[width=\linewidth]{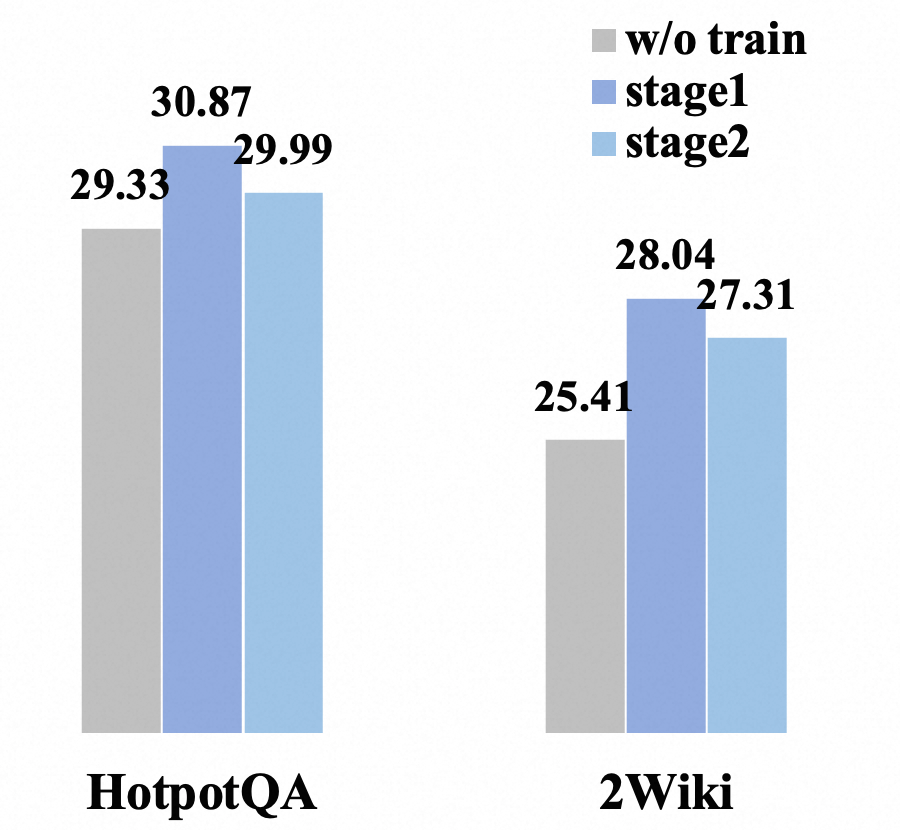}
        \caption{Information utilization ability}
        \label{fig:info}
    \end{minipage}
    \hspace{0.05\linewidth}
    \begin{minipage}[b]{0.45\linewidth}
        \centering
        \includegraphics[width=\linewidth]{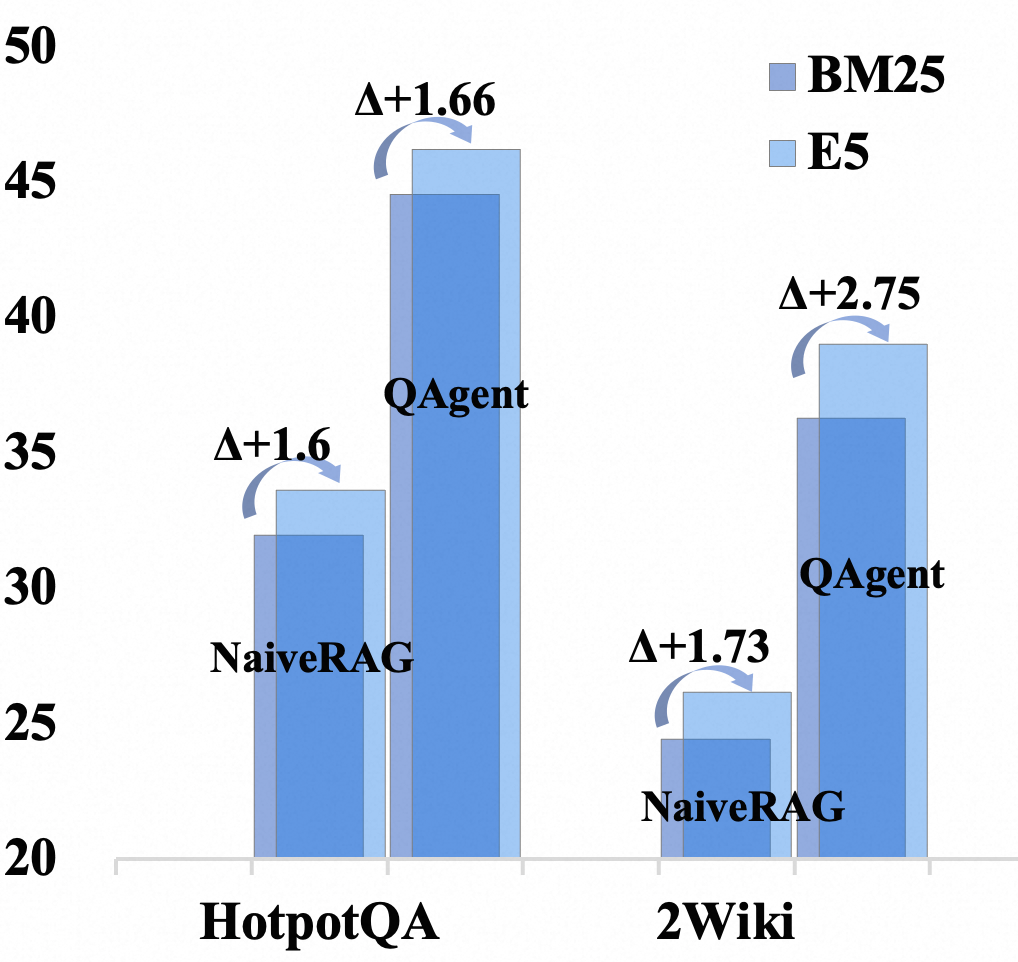}
        \caption{Different retrievers}
        \label{fig:retriever}
    \end{minipage}
\end{figure}

\subsection{Information Utilization Analysis}
\label{sec:utilization}
As analyzed in Section~\ref{sec:Generalized_Train}, as the model training, its information utilization capability improves, and this is a obstacle to the model's information retrieval capability. Here, we analyze the two-stage trained models' ability to utilize the same information. Specifically, we perform naive RAG on the trained models, give them the same top-1 document, and then test the performance of the responses. We found that the end-to-end trained model achieved the highest performance. This is because information utilization capability is the direct goal of training. After generalization training, the model's information utilization capability decreased. We speculate that this is because the optimization goal at this time is no longer focused on information utilization, but gradually focuses on information retrieval capability. This provides new insights into the training of search agents.

\subsection{Generalization to Different Retrievers}
\label{sec:retriever}
In this section, we analyzed the generalization of QAgent. 
As shown in Figure~\ref{fig:retriever}, we used two representative search engines, BM25 and E5.
As expected, using the more powerful search engine, E5, yielded better performance. 
Specifically, the strong retriever enables QAgent to achieve improvements on HotpotQA similar to Naive RAG.
However, powerful search engine yielded a greater performance improvement on 2WikiMHQ compared to Naive RAG. 
This suggests that the QAgent middleware is plug-and-play, compatible with both search engines and generators, and exhibiting high generalization performance.

\begin{figure}[!h]
    \centering
    \includegraphics[width=0.775\linewidth]{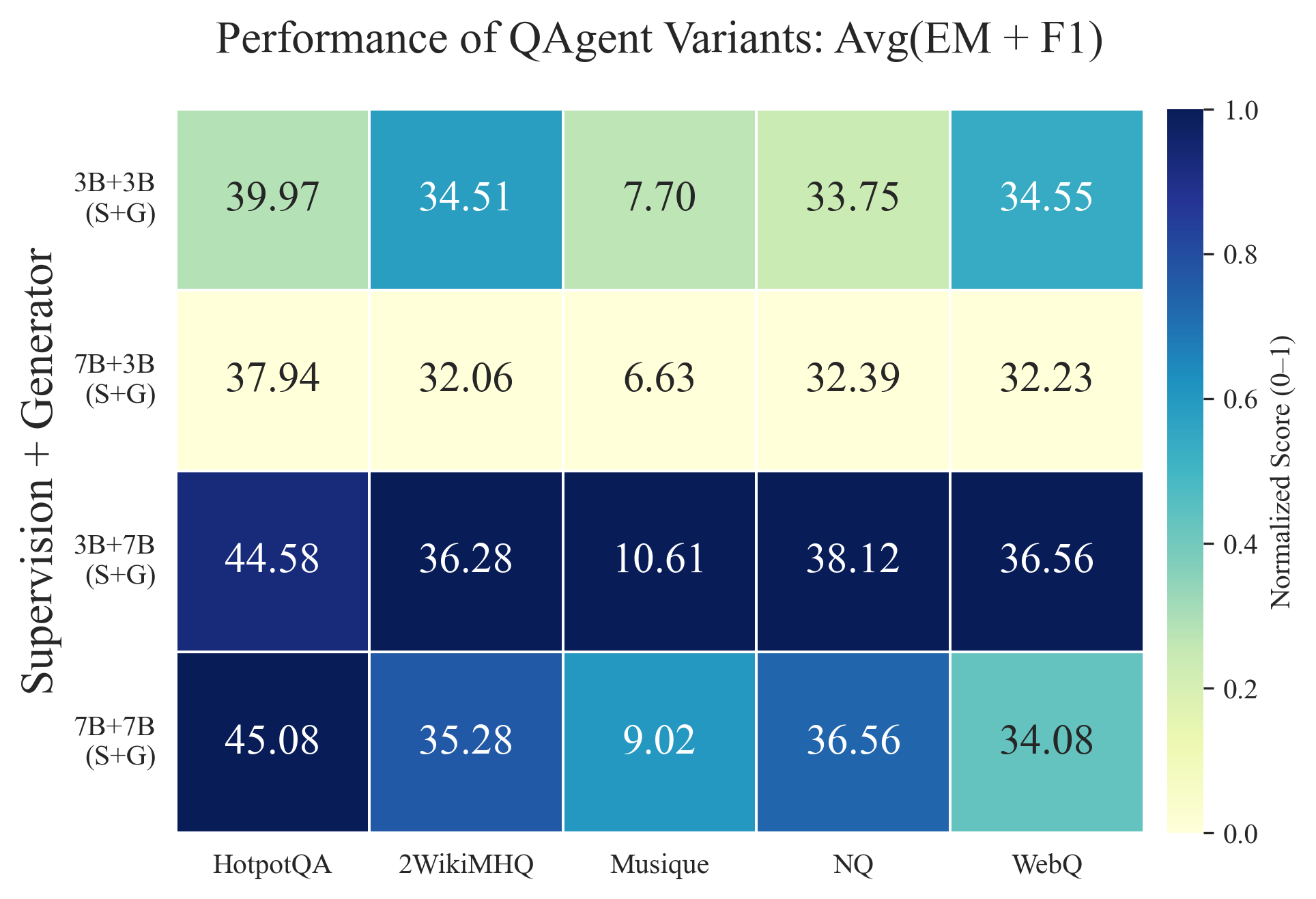}
    \caption{The performance difference between using different generators during training and inference. }
    \label{fig:cross_heatmap}
\end{figure}

\subsection{Different Model Sizes for Supervision}
\label{sec:supervision}
Using the generator's responses to indirectly calculate the reward, rather than relying on end-to-end answer, helps avoid reward hacking that may arise from over-optimizing information utilization. 
However, the impact of different generators for supervision requires further analysis. We experimented with two generator sizes during training: 3B and 7B. As shown in Figure~\ref{fig:cross_heatmap}, training with the 7B generator led to more stable, but 3B achieved better overall performance. We speculate that this results from the 3B model’s inherent instability, which acts as a form of regularization. Specifically, if a smaller generator can produce correct answers, it suggests that the retrieved information is sufficient for accurate reasoning. 

During inference, larger generators consistently yield better performance. This indicates that complex systems benefit from more powerful generators, and a lightweight search agent can be effectively integrated to meet efficiency.

\section{Conclusion}
\label{Conclusion}
In this work, we present QAgent, a unified agentic RAG framework that tackles two key challenges: weak complex query understanding and limited generalization of search agents. QAgent employs a modular, plug and play agent that optimizes queries through iterative reasoning and retrieval, trained with RL to maximize retrieval quality. 
Experiments show QAgent outperforms existing methods on complex QA tasks. 

\section{Limitation}
While QAgent achieves strong performance, several limitations remain. 

\paragraph{Training on larger models.} 
Our motivation is to serve as a submodule of the system. Within this goal, validation on smaller models is paramount. However, whether larger models achieve consistent results remains to be verified.
\paragraph{Failure to control passage diversity.} 
We tried various approaches to passage diversity, including designing repeated negative rewards and adjusting passage overlap, but found that these all led to severe reward hacking, especially when the number of passages is uncertain. Diversity and number are strongly correlated, and the model may overreact by increasing the number of irrelevant passages.
\paragraph{Robustness challenges.} 
The retriever is a factor that influences training and inference, affecting the agent's learning patterns, particularly re-retrieval.
Furthermore, intuitively, query understanding and search engine preferences are relatively consistent. For example, sparse retriever tend to favor keywords. Although we have verified that QAgent achieves robust performance on different search engines, it remains a challenge to directly achieve a query optimization goal that satisfies the preferences of all search engines.

\bibliography{custom}

\begin{thebibliography}{50}
\providecommand{\natexlab}[1]{#1}

\bibitem[{Asai et~al.(2024)Asai, Wu, Wang, Sil, and Hajishirzi}]{asai2024self}
Akari Asai, Zeqiu Wu, Yizhong Wang, Avirup Sil, and Hannaneh Hajishirzi. 2024.
\newblock Self-{RAG}: Learning to retrieve, generate, and critique through self-reflection.
\newblock In \emph{The Twelfth International Conference on Learning Representations}.

\bibitem[{B{\'e}chard and Ayala(2024)}]{bechard2024reducing}
Patrice B{\'e}chard and Orlando~Marquez Ayala. 2024.
\newblock Reducing hallucination in structured outputs via retrieval-augmented generation.
\newblock \emph{arXiv preprint arXiv:2404.08189}.

\bibitem[{Berant et~al.(2013)Berant, Chou, Frostig, and Liang}]{berant2013semantic}
Jonathan Berant, Andrew Chou, Roy Frostig, and Percy Liang. 2013.
\newblock Semantic parsing on freebase from question-answer pairs.
\newblock In \emph{Proceedings of the 2013 conference on empirical methods in natural language processing}, pages 1533--1544.

\bibitem[{Cheng et~al.(2024)Cheng, Li, Li, Zhu, Yin, Shao, Li, Sun, Yan, and Qiu}]{cheng2024unified}
Qinyuan Cheng, Xiaonan Li, Shimin Li, Qin Zhu, Zhangyue Yin, Yunfan Shao, Linyang Li, Tianxiang Sun, Hang Yan, and Xipeng Qiu. 2024.
\newblock Unified active retrieval for retrieval augmented generation.
\newblock In \emph{Findings of the Association for Computational Linguistics: EMNLP 2024}, pages 17153--17166.

\bibitem[{Dong et~al.(2025)Dong, Zhu, Zhang, Wang, Wen, and Dou}]{dong2025understand}
Guanting Dong, Yutao Zhu, Chenghao Zhang, Zechen Wang, Ji-Rong Wen, and Zhicheng Dou. 2025.
\newblock Understand what llm needs: Dual preference alignment for retrieval-augmented generation.
\newblock In \emph{Proceedings of the ACM on Web Conference 2025}, pages 4206--4225.

\bibitem[{Fan et~al.(2024)Fan, Ding, Ning, Wang, Li, Yin, Chua, and Li}]{fan2024survey}
Wenqi Fan, Yujuan Ding, Liangbo Ning, Shijie Wang, Hengyun Li, Dawei Yin, Tat-Seng Chua, and Qing Li. 2024.
\newblock A survey on rag meeting llms: Towards retrieval-augmented large language models.
\newblock In \emph{Proceedings of the 30th ACM SIGKDD conference on knowledge discovery and data mining}, pages 6491--6501.

\bibitem[{Gade et~al.(2025)Gade, Jetcheva, and Trivedi}]{gade2025s}
Anoushka Gade, Jorjeta~G Jetcheva, and Hardi Trivedi. 2025.
\newblock It's about time: Incorporating temporality in retrieval augmented language models.
\newblock In \emph{2025 IEEE Conference on Artificial Intelligence (CAI)}, pages 75--82. IEEE.

\bibitem[{Glass et~al.(2022)Glass, Rossiello, Chowdhury, Naik, Cai, and Gliozzo}]{glass2022re2g}
Michael Glass, Gaetano Rossiello, Md~Faisal~Mahbub Chowdhury, Ankita Naik, Pengshan Cai, and Alfio Gliozzo. 2022.
\newblock Re2g: Retrieve, rerank, generate.
\newblock In \emph{Proceedings of the 2022 Conference of the North American Chapter of the Association for Computational Linguistics: Human Language Technologies}, pages 2701--2715.

\bibitem[{Guo et~al.(2025)Guo, Yang, Zhang, Song, Zhang, Xu, Zhu, Ma, Wang, Bi et~al.}]{guo2025deepseek}
Daya Guo, Dejian Yang, Haowei Zhang, Junxiao Song, Ruoyu Zhang, Runxin Xu, Qihao Zhu, Shirong Ma, Peiyi Wang, Xiao Bi, and 1 others. 2025.
\newblock Deepseek-r1: Incentivizing reasoning capability in llms via reinforcement learning.
\newblock \emph{arXiv preprint arXiv:2501.12948}.

\bibitem[{He et~al.(2025)He, Basulto, and Pan}]{he2025sufficiency}
Jie He, V{\'\i}ctor~Guti{\'e}rrez Basulto, and Jeff~Z Pan. 2025.
\newblock From sufficiency to reflection: Reinforcement-guided thinking quality in retrieval-augmented reasoning for llms.
\newblock \emph{arXiv preprint arXiv:2507.22716}.

\bibitem[{Ho et~al.(2020)Ho, Nguyen, Sugawara, and Aizawa}]{ho2020constructing}
Xanh Ho, Anh-Khoa~Duong Nguyen, Saku Sugawara, and Akiko Aizawa. 2020.
\newblock Constructing a multi-hop qa dataset for comprehensive evaluation of reasoning steps.
\newblock In \emph{Proceedings of the 28th International Conference on Computational Linguistics}, pages 6609--6625.

\bibitem[{Izacard et~al.(2023)Izacard, Lewis, Lomeli, Hosseini, Petroni, Schick, Dwivedi-Yu, Joulin, Riedel, and Grave}]{izacard2023atlas}
Gautier Izacard, Patrick Lewis, Maria Lomeli, Lucas Hosseini, Fabio Petroni, Timo Schick, Jane Dwivedi-Yu, Armand Joulin, Sebastian Riedel, and Edouard Grave. 2023.
\newblock Atlas: Few-shot learning with retrieval augmented language models.
\newblock \emph{Journal of Machine Learning Research}, 24(251):1--43.

\bibitem[{Jaech et~al.(2024)Jaech, Kalai, Lerer, Richardson, El-Kishky, Low, Helyar, Madry, Beutel, Carney et~al.}]{jaech2024openai}
Aaron Jaech, Adam Kalai, Adam Lerer, Adam Richardson, Ahmed El-Kishky, Aiden Low, Alec Helyar, Aleksander Madry, Alex Beutel, Alex Carney, and 1 others. 2024.
\newblock Openai o1 system card.
\newblock \emph{arXiv preprint arXiv:2412.16720}.

\bibitem[{Jeong et~al.(2024)Jeong, Baek, Cho, Hwang, and Park}]{jeong2024adaptive}
Soyeong Jeong, Jinheon Baek, Sukmin Cho, Sung~Ju Hwang, and Jong~C Park. 2024.
\newblock Adaptive-rag: Learning to adapt retrieval-augmented large language models through question complexity.
\newblock In \emph{Proceedings of the 2024 Conference of the North American Chapter of the Association for Computational Linguistics: Human Language Technologies (Volume 1: Long Papers)}, pages 7029--7043.

\bibitem[{Jiang et~al.(2023)Jiang, Wu, Luo, Li, Lin, Yang, and Qiu}]{jiang2023longllmlingua}
Huiqiang Jiang, Qianhui Wu, Xufang Luo, Dongsheng Li, Chin-Yew Lin, Yuqing Yang, and Lili Qiu. 2023.
\newblock Longllmlingua: Accelerating and enhancing llms in long context scenarios via prompt compression.
\newblock \emph{arXiv preprint arXiv:2310.06839}.

\bibitem[{Jiang et~al.(2025{\natexlab{a}})Jiang, Xu, Lin, Xiao, Wang, Sun, and Han}]{jiang2025s3}
Pengcheng Jiang, Xueqiang Xu, Jiacheng Lin, Jinfeng Xiao, Zifeng Wang, Jimeng Sun, and Jiawei Han. 2025{\natexlab{a}}.
\newblock s3: You don't need that much data to train a search agent via rl.
\newblock \emph{arXiv preprint arXiv:2505.14146}.

\bibitem[{Jiang et~al.(2025{\natexlab{b}})Jiang, Zhao, Li, Wang, and Qin}]{jiang2025gainrag}
Yi~Jiang, Sendong Zhao, Jianbo Li, Haochun Wang, and Bing Qin. 2025{\natexlab{b}}.
\newblock {G}ain{RAG}: Preference alignment in retrieval-augmented generation through gain signal synthesis.
\newblock In \emph{Proceedings of the 63rd Annual Meeting of the Association for Computational Linguistics (Volume 1: Long Papers)}, pages 10746--10757. Association for Computational Linguistics.

\bibitem[{Jiang et~al.(2025{\natexlab{c}})Jiang, Sun, Liang, and Zhang}]{jiang2025retrieve}
Zhouyu Jiang, Mengshu Sun, Lei Liang, and Zhiqiang Zhang. 2025{\natexlab{c}}.
\newblock Retrieve, summarize, plan: Advancing multi-hop question answering with an iterative approach.
\newblock In \emph{Companion Proceedings of the ACM on Web Conference 2025}, pages 1677--1686.

\bibitem[{Jin et~al.(2025)Jin, Zeng, Yue, Yoon, Arik, Wang, Zamani, and Han}]{jin2025search}
Bowen Jin, Hansi Zeng, Zhenrui Yue, Jinsung Yoon, Sercan Arik, Dong Wang, Hamed Zamani, and Jiawei Han. 2025.
\newblock Search-r1: Training llms to reason and leverage search engines with reinforcement learning.
\newblock \emph{arXiv preprint arXiv:2503.09516}.

\bibitem[{Kaelbling et~al.(1996)Kaelbling, Littman, and Moore}]{kaelbling1996reinforcement}
Leslie~Pack Kaelbling, Michael~L Littman, and Andrew~W Moore. 1996.
\newblock Reinforcement learning: A survey.
\newblock \emph{Journal of artificial intelligence research}, 4:237--285.

\bibitem[{Karpukhin et~al.(2020)Karpukhin, Oguz, Min, Lewis, Wu, Edunov, Chen, and Yih}]{karpukhin2020dense}
Vladimir Karpukhin, Barlas Oguz, Sewon Min, Patrick~SH Lewis, Ledell Wu, Sergey Edunov, Danqi Chen, and Wen-tau Yih. 2020.
\newblock Dense passage retrieval for open-domain question answering.
\newblock In \emph{EMNLP (1)}, pages 6769--6781.

\bibitem[{Ke et~al.(2024)Ke, Kong, Li, Zhang, Mei, and Bendersky}]{ke2024bridging}
Zixuan Ke, Weize Kong, Cheng Li, Mingyang Zhang, Qiaozhu Mei, and Michael Bendersky. 2024.
\newblock Bridging the preference gap between retrievers and llms.
\newblock In \emph{Proceedings of the 62nd Annual Meeting of the Association for Computational Linguistics (Volume 1: Long Papers)}, pages 10438--10451.

\bibitem[{Kwiatkowski et~al.(2019)Kwiatkowski, Palomaki, Redfield, Collins, Parikh, Alberti, Epstein, Polosukhin, Devlin, Lee et~al.}]{kwiatkowski2019natural}
Tom Kwiatkowski, Jennimaria Palomaki, Olivia Redfield, Michael Collins, Ankur Parikh, Chris Alberti, Danielle Epstein, Illia Polosukhin, Jacob Devlin, Kenton Lee, and 1 others. 2019.
\newblock Natural questions: a benchmark for question answering research.
\newblock \emph{Transactions of the Association for Computational Linguistics}, 7:453--466.

\bibitem[{Lewis et~al.(2020)Lewis, Perez, Piktus, Petroni, Karpukhin, Goyal, K{\"u}ttler, Lewis, Yih, Rockt{\"a}schel et~al.}]{lewis2020retrieval}
Patrick Lewis, Ethan Perez, Aleksandra Piktus, Fabio Petroni, Vladimir Karpukhin, Naman Goyal, Heinrich K{\"u}ttler, Mike Lewis, Wen-tau Yih, Tim Rockt{\"a}schel, and 1 others. 2020.
\newblock Retrieval-augmented generation for knowledge-intensive nlp tasks.
\newblock \emph{Advances in neural information processing systems}, 33:9459--9474.

\bibitem[{Li et~al.(2025{\natexlab{a}})Li, Dong, Jin, Zhang, Zhou, Zhu, Zhang, and Dou}]{li2025search}
Xiaoxi Li, Guanting Dong, Jiajie Jin, Yuyao Zhang, Yujia Zhou, Yutao Zhu, Peitian Zhang, and Zhicheng Dou. 2025{\natexlab{a}}.
\newblock Search-o1: Agentic search-enhanced large reasoning models.
\newblock \emph{arXiv preprint arXiv:2501.05366}.

\bibitem[{Li et~al.(2025{\natexlab{b}})Li, Jin, Zhou, Zhang, Zhang, Zhu, and Dou}]{li2025matching}
Xiaoxi Li, Jiajie Jin, Yujia Zhou, Yuyao Zhang, Peitian Zhang, Yutao Zhu, and Zhicheng Dou. 2025{\natexlab{b}}.
\newblock From matching to generation: A survey on generative information retrieval.
\newblock \emph{ACM Transactions on Information Systems}, 43(3):1--62.

\bibitem[{Lin et~al.(2023)Lin, Chen, Chen, Shi, Lomeli, James, Rodriguez, Kahn, Szilvasy, Lewis et~al.}]{lin2023ra}
Xi~Victoria Lin, Xilun Chen, Mingda Chen, Weijia Shi, Maria Lomeli, Richard James, Pedro Rodriguez, Jacob Kahn, Gergely Szilvasy, Mike Lewis, and 1 others. 2023.
\newblock Ra-dit: Retrieval-augmented dual instruction tuning.
\newblock In \emph{The Twelfth International Conference on Learning Representations}.

\bibitem[{Ma et~al.(2023)Ma, Gong, He, Zhao, and Duan}]{ma2023query}
Xinbei Ma, Yeyun Gong, Pengcheng He, Hai Zhao, and Nan Duan. 2023.
\newblock Query rewriting in retrieval-augmented large language models.
\newblock In \emph{Proceedings of the 2023 Conference on Empirical Methods in Natural Language Processing}, pages 5303--5315.

\bibitem[{Ren et~al.(2025)Ren, Xu, Wang, Li, Ma, and Liu}]{ren2025effective}
Jingyi Ren, Yekun Xu, Xiaolong Wang, Weitao Li, Weizhi Ma, and Yang Liu. 2025.
\newblock Effective and transparent rag: Adaptive-reward reinforcement learning for decision traceability.
\newblock \emph{arXiv preprint arXiv:2505.13258}.

\bibitem[{Robertson et~al.(2009)Robertson, Zaragoza et~al.}]{robertson2009probabilistic}
Stephen Robertson, Hugo Zaragoza, and 1 others. 2009.
\newblock The probabilistic relevance framework: Bm25 and beyond.
\newblock \emph{Foundations and Trends{\textregistered} in Information Retrieval}, 3(4):333--389.

\bibitem[{Sha et~al.(2025)Sha, Cui, and Wang}]{sha2025sem}
Zeyang Sha, Shiwen Cui, and Weiqiang Wang. 2025.
\newblock Sem: Reinforcement learning for search-efficient large language models.
\newblock \emph{arXiv preprint arXiv:2505.07903}.

\bibitem[{Shao et~al.(2024)Shao, Wang, Zhu, Xu, Song, Bi, Zhang, Zhang, Li, Wu et~al.}]{shao2024deepseekmath}
Zhihong Shao, Peiyi Wang, Qihao Zhu, Runxin Xu, Junxiao Song, Xiao Bi, Haowei Zhang, Mingchuan Zhang, YK~Li, Yang Wu, and 1 others. 2024.
\newblock Deepseekmath: Pushing the limits of mathematical reasoning in open language models.
\newblock \emph{arXiv preprint arXiv:2402.03300}.

\bibitem[{Song et~al.(2025)Song, Jiang, Min, Chen, Chen, Zhao, Fang, and Wen}]{song2025r1}
Huatong Song, Jinhao Jiang, Yingqian Min, Jie Chen, Zhipeng Chen, Wayne~Xin Zhao, Lei Fang, and Ji-Rong Wen. 2025.
\newblock R1-searcher: Incentivizing the search capability in llms via reinforcement learning.
\newblock \emph{arXiv preprint arXiv:2503.05592}.

\bibitem[{Sun et~al.(2025)Sun, Qiao, Guo, Fan, Hou, Jiang, Xie, Zhang, Huang, and Zhou}]{sun2025zerosearch}
Hao Sun, Zile Qiao, Jiayan Guo, Xuanbo Fan, Yingyan Hou, Yong Jiang, Pengjun Xie, Yan Zhang, Fei Huang, and Jingren Zhou. 2025.
\newblock Zerosearch: Incentivize the search capability of llms without searching.
\newblock \emph{arXiv preprint arXiv:2505.04588}.

\bibitem[{Trivedi et~al.(2022)Trivedi, Balasubramanian, Khot, and Sabharwal}]{trivedi2022musique}
Harsh Trivedi, Niranjan Balasubramanian, Tushar Khot, and Ashish Sabharwal. 2022.
\newblock Musique: Multihop questions via single-hop question composition.
\newblock \emph{Transactions of the Association for Computational Linguistics}, 10:539--554.

\bibitem[{Trivedi et~al.(2023)Trivedi, Balasubramanian, Khot, and Sabharwal}]{trivedi2023interleaving}
Harsh Trivedi, Niranjan Balasubramanian, Tushar Khot, and Ashish Sabharwal. 2023.
\newblock Interleaving retrieval with chain-of-thought reasoning for knowledge-intensive multi-step questions.
\newblock In \emph{Proceedings of the 61st Annual Meeting of the Association for Computational Linguistics (Volume 1: Long Papers)}, pages 10014--10037.

\bibitem[{Wang et~al.(2023{\natexlab{a}})Wang, Yang, and Wei}]{wang2023query2doc}
Liang Wang, Nan Yang, and Furu Wei. 2023{\natexlab{a}}.
\newblock Query2doc: Query expansion with large language models.
\newblock In \emph{Proceedings of the 2023 Conference on Empirical Methods in Natural Language Processing}, pages 9414--9423.

\bibitem[{Wang et~al.(2023{\natexlab{b}})Wang, Li, Sun, and Liu}]{wang2023self}
Yile Wang, Peng Li, Maosong Sun, and Yang Liu. 2023{\natexlab{b}}.
\newblock Self-knowledge guided retrieval augmentation for large language models.
\newblock In \emph{Findings of the Association for Computational Linguistics: EMNLP 2023}, pages 10303--10315.

\bibitem[{Wang et~al.(2025)Wang, Zheng, An, Ouyang, Cai, Wang, and Wu}]{wang2025stepsearch}
Ziliang Wang, Xuhui Zheng, Kang An, Cijun Ouyang, Jialu Cai, Yuhang Wang, and Yichao Wu. 2025.
\newblock Stepsearch: Igniting llms search ability via step-wise proximal policy optimization.
\newblock \emph{arXiv preprint arXiv:2505.15107}.

\bibitem[{Wei et~al.(2025)Wei, Chen, and Meng}]{wei2025instructrag}
Zhepei Wei, Wei-Lin Chen, and Yu~Meng. 2025.
\newblock Instruct{RAG}: Instructing retrieval-augmented generation via self-synthesized rationales.
\newblock In \emph{The Thirteenth International Conference on Learning Representations}.

\bibitem[{Xiao et~al.(2024)Xiao, Liu, Zhang, Muennighoff, Lian, and Nie}]{xiao2024c}
Shitao Xiao, Zheng Liu, Peitian Zhang, Niklas Muennighoff, Defu Lian, and Jian-Yun Nie. 2024.
\newblock C-pack: Packed resources for general chinese embeddings.
\newblock In \emph{Proceedings of the 47th international ACM SIGIR conference on research and development in information retrieval}, pages 641--649.

\bibitem[{Xu et~al.(2024{\natexlab{a}})Xu, Shi, and Choi}]{xu2024recomp}
Fangyuan Xu, Weijia Shi, and Eunsol Choi. 2024{\natexlab{a}}.
\newblock Recomp: Improving retrieval-augmented lms with compression and selective augmentation.
\newblock In \emph{12th International Conference on Learning Representations, ICLR 2024}.

\bibitem[{Xu et~al.(2024{\natexlab{b}})Xu, Pang, Shen, Cheng, and Chua}]{xu2024search}
Shicheng Xu, Liang Pang, Huawei Shen, Xueqi Cheng, and Tat-Seng Chua. 2024{\natexlab{b}}.
\newblock Search-in-the-chain: Interactively enhancing large language models with search for knowledge-intensive tasks.
\newblock In \emph{Proceedings of the ACM Web Conference 2024}, pages 1362--1373.

\bibitem[{Yang et~al.(2025)Yang, Yang, Zhang, Hui, Zheng, Yu, Li, Liu, Huang, Wei, Lin, Yang, Tu, Zhang, Yang, Yang, Zhou, Lin, Dang, Lu, Bao, Yang, Yu, Li, Xue, Zhang, Zhu, Men, Lin, Li, Tang, Xia, Ren, Ren, Fan, Su, Zhang, Wan, Liu, Cui, Zhang, and Qiu}]{qwen2025qwen25technicalreport}
An~Yang, Baosong Yang, Beichen Zhang, Binyuan Hui, Bo~Zheng, Bowen Yu, Chengyuan Li, Dayiheng Liu, Fei Huang, Haoran Wei, Huan Lin, Jian Yang, Jianhong Tu, Jianwei Zhang, Jianxin Yang, Jiaxi Yang, Jingren Zhou, Junyang Lin, Kai Dang, and 23 others. 2025.
\newblock \href {https://arxiv.org/abs/2412.15115} {Qwen2.5 technical report}.
\newblock \emph{Preprint}, arXiv:2412.15115.

\bibitem[{Yang et~al.(2018)Yang, Qi, Zhang, Bengio, Cohen, Salakhutdinov, and Manning}]{yang2018hotpotqa}
Zhilin Yang, Peng Qi, Saizheng Zhang, Yoshua Bengio, William Cohen, Ruslan Salakhutdinov, and Christopher~D Manning. 2018.
\newblock Hotpotqa: A dataset for diverse, explainable multi-hop question answering.
\newblock In \emph{Proceedings of the 2018 Conference on Empirical Methods in Natural Language Processing}, pages 2369--2380.

\bibitem[{Zhao et~al.(2024{\natexlab{a}})Zhao, Zhang, Yu, Wang, Geng, Fu, Yang, Zhang, Jiang, and Cui}]{zhao2024retrieval1}
Penghao Zhao, Hailin Zhang, Qinhan Yu, Zhengren Wang, Yunteng Geng, Fangcheng Fu, Ling Yang, Wentao Zhang, Jie Jiang, and Bin Cui. 2024{\natexlab{a}}.
\newblock Retrieval-augmented generation for ai-generated content: A survey.
\newblock \emph{arXiv preprint arXiv:2402.19473}.

\bibitem[{Zhao et~al.(2024{\natexlab{b}})Zhao, Yang, Wang, He, Qiu, and Qiu}]{zhao2024retrieval2}
Siyun Zhao, Yuqing Yang, Zilong Wang, Zhiyuan He, Luna~K Qiu, and Lili Qiu. 2024{\natexlab{b}}.
\newblock Retrieval augmented generation (rag) and beyond: A comprehensive survey on how to make your llms use external data more wisely.
\newblock \emph{arXiv preprint arXiv:2409.14924}.

\bibitem[{Zheng et~al.(2025)Zheng, Fu, Hu, Cai, Ye, Lu, and Liu}]{zheng2025deepresearcher}
Yuxiang Zheng, Dayuan Fu, Xiangkun Hu, Xiaojie Cai, Lyumanshan Ye, Pengrui Lu, and Pengfei Liu. 2025.
\newblock Deepresearcher: Scaling deep research via reinforcement learning in real-world environments.
\newblock \emph{arXiv preprint arXiv:2504.03160}.

\bibitem[{Zhou et~al.(2024{\natexlab{a}})Zhou, Liu, Li, Jin, Qian, Liu, Li, Dou, Ho, and Yu}]{zhou2024trustworthiness}
Yujia Zhou, Yan Liu, Xiaoxi Li, Jiajie Jin, Hongjin Qian, Zheng Liu, Chaozhuo Li, Zhicheng Dou, Tsung-Yi Ho, and Philip~S Yu. 2024{\natexlab{a}}.
\newblock Trustworthiness in retrieval-augmented generation systems: A survey.
\newblock \emph{arXiv preprint arXiv:2409.10102}.

\bibitem[{Zhou et~al.(2024{\natexlab{b}})Zhou, Liu, Jin, Nie, and Dou}]{zhou2024metacognitive}
Yujia Zhou, Zheng Liu, Jiajie Jin, Jian-Yun Nie, and Zhicheng Dou. 2024{\natexlab{b}}.
\newblock Metacognitive retrieval-augmented large language models.
\newblock In \emph{Proceedings of the ACM Web Conference 2024}, pages 1453--1463.

\end{thebibliography}

\appendix


\section{Dataset}
\label{appx:dataset}
Here, we provide a detailed description of the datasets we used.

\textbf{HotpotQA}\citep{yang2018hotpotqa} and \textbf{2WikiMultiHopQA}\citep{ho2020constructing}: Both datasets are multi-hop question answering datasets based on Wikipedia. Given the cost constraints of the experiment, we used a subsample of the \citet{trivedi2023interleaving} dataset,
which was obtained by extracting 500 questions from the validation set of each dataset.

\textbf{Musique}\citep{trivedi2022musique}: By iteratively selecting composable single-hop question pairs, we created questions spanning 2-4 hops, which are generally more difficult.

\textbf{WebQuestions}\citep{berant2013semantic}: Constructed from questions posed by the Google Suggest API, where the answers are specific entities listed in Freebase.

\textbf{NaturalQA}\citep{kwiatkowski2019natural}: A dataset designed to support comprehensive question answering systems. It contains questions from real Google search queries, and the corresponding answers are text snippets from Wikipedia articles, carefully identified by human annotators.

\section{baselines}
\label{appx:baselines}
We compare QAgent with different baselines: 

\noindent\textbf{Vanilla}: Direct answering without retrieval.

\textbf{Naive RAG}: Predicting an answer in the ``retrieval-then-reading'' paradigm. 

\textbf{Search-o1}: Using multiple rounds of interaction but without RL training.

\textbf{Search-R1}: Using multiple rounds of interaction and joint retrieval and generation model with RL training.

\textbf{zero-search}: Training a LLM as retriever can reduce the cost and enhance the robustness to the retriever.

\textbf{Rerank}: Follows the ``retrieve-rearrange-generate'' paradigm and is the most widely used middleware (this does not conflict with QAgent). 

When as the search agent component, we extract the retrieved documents from the model inference trajectory and input them into the frozen generators.

\section{Implementation Details}
We experimented with Qwen-2.5-3B (Instruct)\citep{qwen2025qwen25technicalreport}. For retrieval, we used the 2018 Wikipedia dump\citep{karpukhin2020dense} as the knowledge source and BM25\citep{robertson2009probabilistic} as the retriever. To ensure a fair comparison, we set the number of retrieved passages to 5 for all retrieval-based methods and 1 for iterative methods.
For training, we followed Search-R1, merging the training sets of NQ\citep{kwiatkowski2019natural} and HotpotQA\citep{yang2018hotpotqa} to form a unified dataset.

For GRPO training, we set the learning rate of the policy LLM to 1e-6. The KL divergence regularization coefficient $\beta$ and the clipping rate $\epsilon$ were set to 0.001 and 0.2, respectively. 
We sampled 5 responses for each prompt. The maximum sequence length was set to 8192 tokens, the maximum response length was 512. 
To accelerate sampling, we use vLLM\footnote{https://docs.vllm.ai/en/latest/} as the inference acceleration framework. We set the rollout temperature and top-p value of our vLLM-based rollout to 1.0.
A purely online algorithm was used, meaning that the vLLM weights used for sampling were updated after each weight update. 
Training is performed on a single node equipped with 8 GPUs, four of which are used for running rollout, three for training, and one for deploying the reference model. 
For training, we used Liger-Kernel\footnote{https://github.com/linkedin/Liger-Kernel} to optimize training efficiency.
The actual batch size for a single GPU is 1, the gradient accumulation is set to 64, and training is performed until 1400 actual update steps, with a warm-up step number of 225. In addition, we took a checkpoint of 1120 steps for stage 1 training, and a checkpoint of 1400 steps for stage 2. 
During the evaluation, we used the vllm framework and performed inference on a single GPU.

\section{Training}
\begin{figure}[!h]
    \centering
    \includegraphics[width=0.7\linewidth]{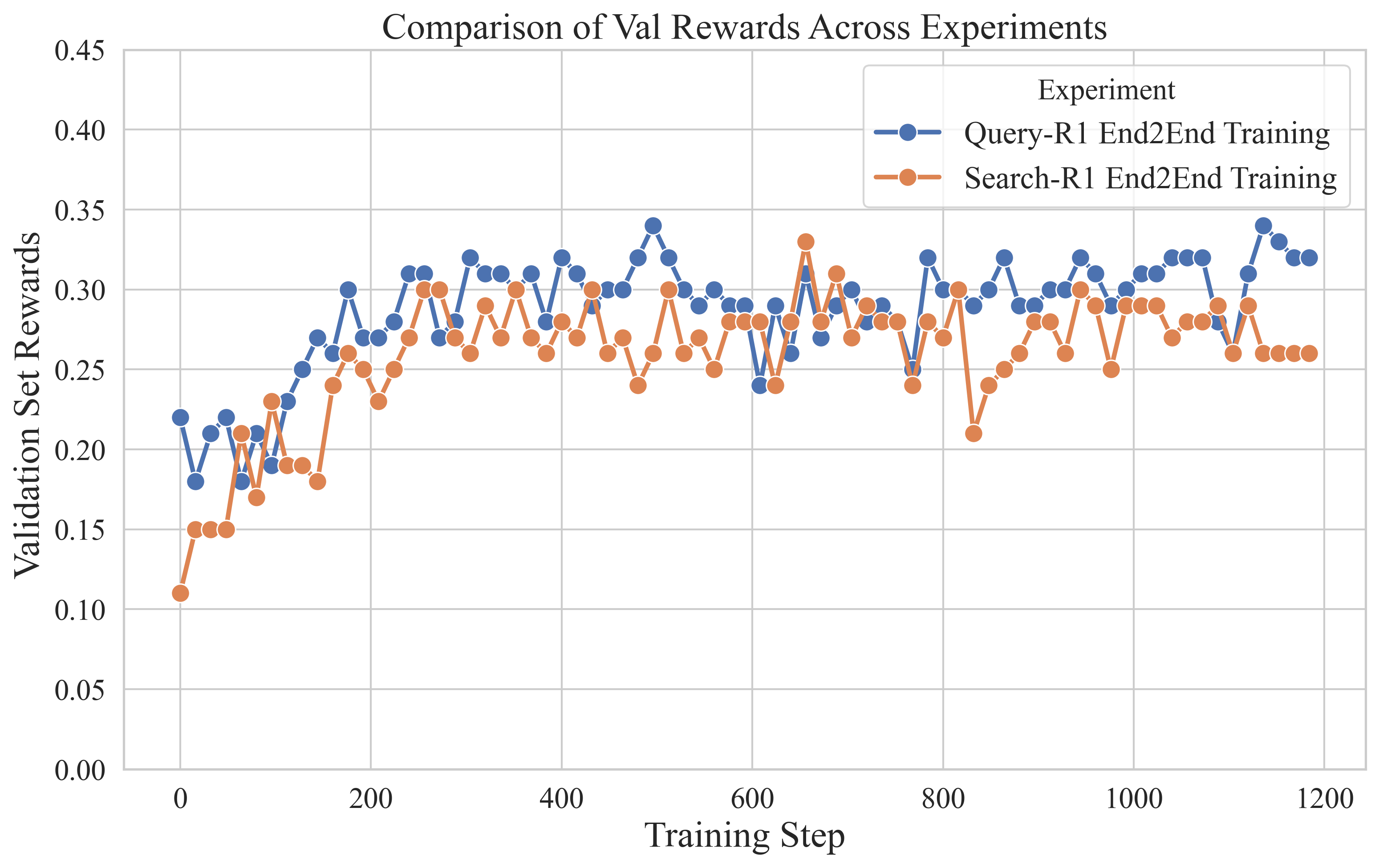}
    \caption{Illustration of the reward increasing during end-to-end training.}
    \label{fig:stage1}
\end{figure}

\begin{figure}[!h]
    \centering
    \includegraphics[width=0.7\linewidth]{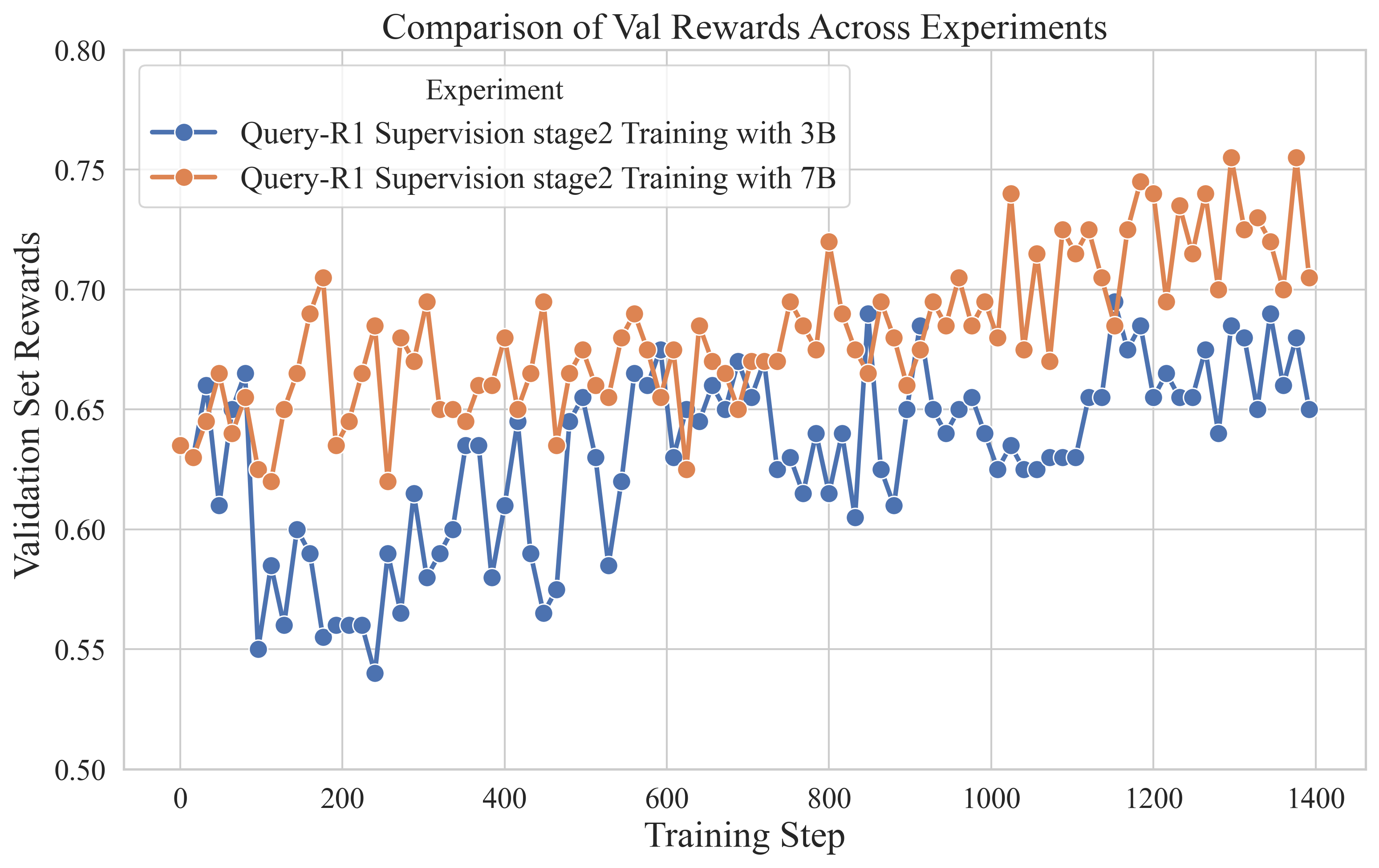}
    \caption{Illustration of the reward increasing during stage2 training. Note that the upper limits of the rewards is different.}
    \label{fig:stage2}
\end{figure}

Figure~\ref{fig:stage1} shows the stability of the reward increase. Compared with Search-R1, our method has a higher starting point and a higher upper limit of convergence. 
As shown in Figure~\ref{fig:stage2}, In the second stage of training, we used different frozen generators for supervision, and we can see that the reward steadily increased. 
Using 7B supervision is more stable and has higher rewards, because the larger generator has richer internal knowledge. However, from the experimental evaluation results, 3B has slightly better generalization.

\section{Experimental Results}
We present here the complete experimental results of Section~\ref{sec:supervision}, as shown in Table~\ref{tab:supervison}.

\begin{table*}[!h]
\centering
\resizebox{0.975\textwidth}{!}{
\begin{tabular}{lcccccccccccc}
\toprule
\multicolumn{1}{c}{} & \multicolumn{2}{c}{HotpotQA} & \multicolumn{2}{c}{2WikiMHQ$^\dagger$} & \multicolumn{2}{c}{MuSique$^\dagger$} & \multicolumn{2}{c}{NaturalQA} & \multicolumn{2}{c}{WebQuestion$^\dagger$} & \multicolumn{2}{c}{Average} \\
\multicolumn{1}{c}{\multirow{-2}{*}{Method}} & EM    & F1    & EM    & F1   & EM   & F1   & EM     & F1    & EM     & F1     & EM   & F1           \\
\midrule
\multicolumn{13}{c}{\cellcolor[HTML]{EFEFEF}Generator 3B}                                                                                                                                                                                  \\
End-to-End   & 36.20 & 41.21 & 30.60 & 33.19        & 5.40 & 10.24        & 30.03  & 30.04 & 30.91  & 30.20  & 26.63        & 28.98        \\
Supervision: 3B   & 37.00 & 42.94 & 33.80 & 35.22        & 5.20 & 10.20        & 33.82  & 33.68 & 36.32  & 32.78  & 29.23        & 30.96        \\
Supervision: 7B   & 34.80 & 41.07 & 31.40 & 32.72        & 4.20 & 9.07 & 32.49  & 32.30 & 32.73  & 31.74  & 27.12        & 29.38        \\
\midrule
\multicolumn{13}{c}{\cellcolor[HTML]{EFEFEF}Generator 7B}                                                                                                                                                                                  \\
End-to-End   & 40.40 & 45.84 & 35.20 & 37.65        & 7.40 & 13.60        & 33.52  & 34.93 & 32.58  & 33.83  & 29.82        & 33.17        \\
Supervision: 3B   & 42.40 & 46.75 & 35.80 & 36.76        & 7.40 & 13.81        & 37.37  & 38.87 & 36.71  & 36.41  & 31.94        & 34.52        \\
Supervision: 7B   & 42.00 & 48.16 & 33.80 & 36.77        & 6.00 & 12.04        & 35.87  & 37.26 & 33.51  & 34.65  & 30.24        & 33.78       \\
\bottomrule
\end{tabular}
}
\caption{Full experimental results using different models as supervision.}
\label{tab:supervison}
\end{table*}

\begin{table*}[!h]
\centering
\begin{tabularx}{0.995\linewidth}{X}
\toprule
Search for information to answer the given question.
You can search as many times as needed if you find you lack some knowledge. 
You will go through a loop of: \\
``\textless plan\textgreater xxx\textless/plan\textgreater \\
\textless search \textgreater  xxx\textless/search\textgreater \\
\textless information\textgreater xxx\textless /information\textgreater \\
\textless reflection\textgreater xxx \textless/reflection\textgreater  \\
\textless plan\textgreater xxx\textless/plan\textgreater  (if not complete)\\
\quad ... \\
\textless reflection\textgreater xxx \textless/reflection\textgreater \\
\textless answer\textgreater xxx\textless /answer\textgreater ''. \\
You must conduct planning inside \textless plan\textgreater  and \textless/plan\textgreater  first every time you call a search engine. \\
After planing, you can call a search engine to search multiple queries (no more than 3) by \\
\textless search\textgreater \\
\quad \textless query\textgreater query1\textless/query\textgreater \\
\quad \textless query\textgreater query2\textless/query\textgreater \\
\quad ... \\
\quad \textless query\textgreater queryk\textless/query\textgreater \\
\textless/search\textgreater, \\
and it will return the searched results between \textless information\textgreater  and \textless /information\textgreater . \\

After getting information, you must conduct a reflection on the information and place your reflection between the \textless reflection\textgreater  and \textless/reflection\textgreater tags. \\
Note that you must plan within \textless plan\textgreater  and \textless/plan\textgreater  before searching, and reflect within \textless reflect\textgreater and \textless/reflect\textgreater after receiving information.\\
Note that each query must be enclosed between \textless query\textgreater  and \textless/query\textgreater , and all queries must be placed between \textless search\textgreater  and \textless/search\textgreater , such as \textless search\textgreater \verb|\n| \textless query\textgreater query1\textless/query\textgreater \verb|\n| \textless query\textgreater query2\textless/query\textgreater \verb|\n|  ... \verb|\n| \textless query\textgreater queryk\textless/query\textgreater \verb|\n| \textless/search\textgreater. \\
If the task is not yet complete, begin a new \textless plan\textgreater. \\
If you find no further external knowledge needed, you can directly provide the answer inside \textless answer\textgreater  and \textless/answer\textgreater  without detailed illustrations. \\
For example, \textless answer\textgreater \verb|\n| xxx\verb|\n| \textless/answer\textgreater . \\
The answer, ``xxx'', should be a few short words. \\
Question: \{$question$\}. 

\\
\bottomrule
\end{tabularx}
\caption{Template prompt used in experiments.}
\label{tab:prompt}
\end{table*}

\section{Template Prompt}
We present here the instruction prompt in Table~\ref{tab:prompt}.

\section{Case Study}
\begin{figure*}[!ht]
    \centering
    \includegraphics[width=0.8975\linewidth]{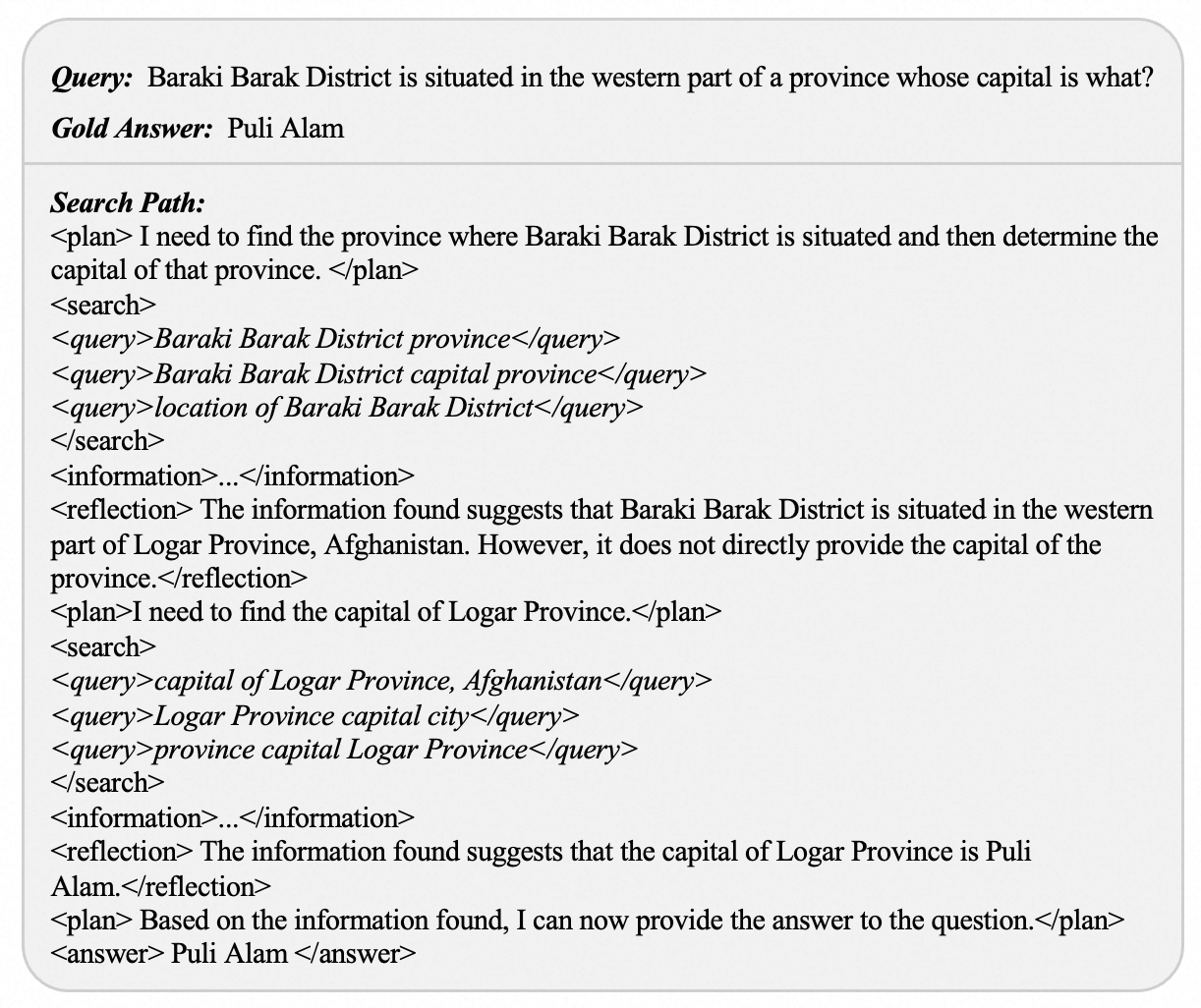}
    \caption{
    A case of QAgent. 
    }
    \label{fig:case_study}
\end{figure*}

A case of QAgent is shown in Figure~\ref{fig:case_study}.

\end{document}